\documentclass[manuscript,screen]{acmart}

\usepackage{longtable}
\usepackage{tikz}
\usepackage{multirow}
\usepackage{array}
\usepackage{tabularx}
\usepackage{booktabs}
\usepackage{makecell}

\usepackage{setspace}
\usepackage{placeins}
\usepackage{afterpage}
\usepackage[utf8]{inputenc}
\usepackage{graphicx}
\usepackage{balance}
\usepackage{amsthm}
\usepackage{color}
\usepackage{caption}
\usepackage{amsmath}
\usepackage{tabu}
\usepackage{array}
\usepackage{booktabs}
\usepackage{threeparttable}
\usepackage{relsize}
\usepackage{physics}
\usepackage{balance}

\usepackage{algorithm}
\usepackage[noend]{algpseudocode}

\newlength{\maxwidth}
\newcommand{\algalign}[2]
{\makebox[\maxwidth][r]{$#1{}$}${}#2$}

\usepackage{float}
\usepackage{stfloats}
\usepackage{lipsum}
\usepackage[english]{babel}

\usepackage{comment}

\usepackage[T1]{fontenc}
\usepackage{mwe}    
\usepackage{subfig}
\usepackage{tabularray}
\usepackage{makecell}

\usepackage{titlesec}
\titleformat{\paragraph}
{\normalfont\normalsize\bfseries}{\theparagraph}{0.5em}{}
\titlespacing*{\paragraph}
{0pt}{2.25ex plus 1ex minus .2ex}{0.5ex plus .2ex}

\titleformat{\subparagraph}
{\normalfont\normalsize\bfseries}{\thesubparagraph}{0.5em}{}
\titlespacing*{\subparagraph}
{0pt}{2.25ex plus 1ex minus .2ex}{0.5ex plus .2ex}

\makeatletter

\renewcommand\subsubsection{\@startsection{subsubsection}{3}
  \z@{.5\linespacing\@plus.7\linespacing}{.5\linespacing}%
  {\normalfont\bfseries}}
\makeatother

\let\oldnl\nl
\newcommand{\nonl}{\renewcommand{\nl}{\let\nl\oldnl}}

\theoremstyle{definition}

\makeatletter
\def\thm@space@setup{%
  \thm@preskip=0pt
  \thm@postskip=0pt
}
\makeatother

\theoremstyle{definition}
\newtheorem{defn}{Definition}[section]

\usepackage{titlesec}
\titlespacing{\section}{0pt}{\parskip}{-\parskip}

\begin{document}

\title{Holistic Survey of Privacy and Fairness in Machine Learning}
\author{Sina Shaham}
\affiliation{%
 \institution{University of Southern California}
 \city{Los Angeles}
  \state{California} 
  \country{USA}
  }
\email{sshaham@usc.edu}

\author{Arash Hajisafi\textsuperscript{*}}
\affiliation{%
 \institution{University of Southern California}
 \city{Los Angeles}
  \state{California} 
  \country{USA}
  }
\email{hajisafi@usc.edu}

\author{Minh K. Quan\textsuperscript{*}}
\affiliation{%
 \institution{Deakin University}
  \state{VIC} 
  \country{Australia}
  }
\email{m.quan@deakin.edu.au}

\author{Dinh C. Nguyen}
\affiliation{%
 \institution{University of Alabama in Huntsville}
 \city{Huntsville}
  \state{Alabama} 
  \country{USA}
  }
\email{Dinh.Nguyen@uah.edu}

\author{Bhaskar Krishnamachari}
\affiliation{%
  \institution{University of Southern California}
  \city{Los Angeles} 
 \state{California} 
  \country{USA}
}
\email{bkrishna@usc.edu}

\author{Charith Peris}
\affiliation{%
  \institution{Amazon}
  \city{Cambridge} 
 \state{MA} 
  \country{USA}
}
\email{perisc@amazon.com}

\author{Gabriel Ghinita}
\affiliation{%
  \institution{Hamad Bin Khalifa University}
  \city{Doha} 
  \country{Qatar}
}
\email{gghinita@hbku.edu.qa}

\author{Cyrus Shahabi}
\affiliation{%
  \institution{University of Southern California}
  \city{Los Angeles} 
 \state{California} 
  \country{USA}
}
\email{shahabi@usc.edu}

\author{Pubudu N. Pathirana}
\affiliation{%
 \institution{Deakin University}
  \state{VIC} 
  \country{Australia}
  }
\email{pubudu.pathirana@deakin.edu.au}

\renewcommand\thefootnote{*}
\footnotetext{These authors contributed equally to this work.}

\renewcommand{\shortauthors}{Sina Shaham, et al.}

\begin{abstract}
Privacy and fairness are two crucial pillars of responsible Artificial Intelligence (AI) and trustworthy Machine Learning (ML). Each objective has been independently studied in the literature with the aim of reducing utility loss in achieving them. Despite the significant interest attracted from both academia and industry, there remains an immediate demand for more in-depth research to unravel how these two objectives can be simultaneously integrated into ML models. As opposed to well-accepted trade-offs, i.e., privacy-utility and fairness-utility, the interrelation between privacy and fairness is not well-understood. While some works suggest a trade-off between the two objective functions, there are others that demonstrate the alignment of these functions in certain scenarios. To fill this research gap, we provide a thorough review of privacy and fairness in ML, including supervised, unsupervised, semi-supervised, and reinforcement learning. After examining and consolidating the literature on both objectives, we present a holistic survey on the impact of privacy on fairness, the impact of fairness on privacy, existing architectures, their interaction in application domains, and algorithms that aim to achieve both objectives while minimizing the utility sacrificed. Finally, we identify research challenges in achieving privacy and fairness concurrently in ML, particularly focusing on large language models.
\end{abstract}


\maketitle

\section{Introduction}

The rapid expansion of big data, along with the rise in computational resources, have allowed for remarkable gains in the capabilities of ML algorithms, igniting a competitive landscape in this field. These algorithms, initially devised by humans, now actively participate in decision-making and policy formation for the same people who created them. The advantages of these algorithms are vast, as they enhance efficiency, accuracy, and speed in various domains. They contribute to improved legal outcomes~\cite{surden2014machine}, streamlined lending and hiring processes~\cite{chalfin2016productivity}, and optimized allocation of resources and benefits~\cite{hussain2020machine}. Harnessing the power of ML to develop equitable and efficient systems can catalyze both social and economic progress.



{\bf Trustworthy ML.} The initial belief that more data would result in better decision-making in the world of ML was quickly shattered as it became clear that accurate algorithms alone are not enough to make responsible decisions~\cite{dwork2012fairness,chen2022fairness,mehrabi2021survey}. The significance of trustworthiness in ML can be explored by making an analogy with the stages of human development. Consider a child who inherits characteristics from their parents – this is akin to the initial model selection in ML, taking into account the mathematical limitations inherent in the chosen structure. As the child matures, they absorb crucial knowledge during their formative years - comparable to an ML model being trained with carefully selected datasets. The child's interaction with their socio-economic environment shapes their behavior and choices, just as an ML model's responses are influenced by the dataset it interacts with and the feedback it receives. The child experiences both opportunities and limitations in society, just as an ML model's functionality is affected by the boundaries of its mathematical design and the quality of its datasets. Ultimately, the child matures into an individual whose ethical decisions impact those around them, much like an ML model that must make responsible decisions affecting real-world outcomes. To develop a trustworthy ML pipeline, each element in the learning cycle, like each stage in a child's life, carries shared responsibility. In this research, we focus on understanding and integrating the two main pillars of trustworthy ML: Privacy and Fairness.

{\bf Privacy.} Information privacy refers to an individual's right to maintain a certain level of control over how their personal data is gathered and utilized~\cite{solove2008understanding}. Take, for instance, a photo shared by an individual on social media platforms, intended solely for communication and social interaction. Even basic data mining techniques can extract sensitive information from this image, which could be exploited by malicious attackers. Aspects like ornaments, background, and facial features may inadvertently disclose the individual's religion, geographical location, gender, race, or other sensitive details. This raises the question of how much control users have over their own data. Expanding this concept to the vast quantities of data utilized in modern ML models highlights the importance of privacy for ensuring trustworthy ML. Incidents like the 2020 Facebook scandal~\cite{FinancialTimes2020} and the Edward Snowden revelations~\cite{ForbesTechCouncil2023} underscore the critical nature of user data privacy in the context of ML.

{\bf Fairness.} From a different perspective, while privacy deals with the extent of control over data, fairness aims to ensure that the revealed user information is handled fairly and equitably. The philosophical notions of fairness have existed for centuries~\cite{anderson2004pursuit}; however, with the rapid growth of ML, algorithmic fairness and its application to ML have emerged as some of the most critical challenges of the decade. Unfortunately, models intended to intelligently avoid errors and biases in decision-making have themselves become sources of bias and discrimination within society. Various forms of unfairness in ML have raised concerns, including racial biases in criminal justice systems~\cite{angwin_larson_kirchner_mattu_2016} and disparities in employment~\cite{raghavan2020mitigating} and loan approval processes~\cite{kozodoi2022fairness}. The entire life-cycle of an ML model – encompassing input data, modeling, evaluation, and feedback – is vulnerable to both external and inherent biases, leading to unjust outcomes.  Compounding the issue is the tendency of the pipeline's life-cycle to amplify biases due to oversimplification and assumptions made throughout the process. Moreover, unlike the concept of privacy, for which there are well-defined and accepted metrics, the large number of varied and often conflicting definitions of fairness presents a significant challenge in establishing trustworthy ML systems.

{\bf Privacy vs Fairness.} Investigations into privacy and fairness have often been carried out separately, without a holistic comprehension of how these two goals intertwine. Although the trade-offs between privacy and utility, as well as fairness and utility, are well-established, the complex relationship between these objectives remains less clear. Several studies such as \cite{bagdasaryan2019differential} and \cite{chen2022fairness}, indicate the presence of trade-offs, while others, like \cite{khalili2021improving} and \cite{pannekoek2021investigating}, consider them to be in harmony. Given the lack of studies elucidating their interconnection, there is an urgent need for more research to uncover the link between these two goals, ultimately paving the way for truly responsible ML models.

{\bf Motivation.} Pursuing privacy and fairness as separate objectives may appear intuitive, yet this approach is fraught with significant issues. First, engineers and researchers often find that the attainment of even one of these goals can significantly impact a model's performance, necessitating careful alignment of both objectives. Second, research exploring how the achievement of one goal influences the other remains limited. This knowledge gap introduces uncertainty concerning the model's reliability. As a result, our goal is to bridge this divide between privacy and fairness, traditionally pursued as independent objectives. We aim to establish a foundation for more advanced techniques facilitating their concurrent implementation, honoring these elements as the two primary pillars of trustworthy ML models.

{\bf Contribution.} In this comprehensive survey, we present an in-depth examination of the main concepts in privacy and fairness by analyzing nearly $200$ recent studies in the field. We explore these approaches across four primary aspects of ML, namely, Supervised Learning (SL), Unsupervised Learning (UL), Semi-Supervised Learning (SSL), and Reinforcement Learning (RL), with the aim of consolidating terminology and ideas. For instance, we collate and explain $15$ distinct fairness notions to facilitate a better understanding of the principles. By establishing a solid comprehension of privacy and fairness across various ML techniques, we offer an extensive review of existing research on architectures designed to meet these goals, the interplay between the objectives, their concurrent implementation, and ultimately, their manifestation in several applications. Moreover, we identify several key unresolved questions and challenges in understanding two objective functions, from large language models to the disparate impact of privacy-preserving methods in ML.

The rest of this survey is organized as follows. A detailed examination of privacy within the realm of ML is presented in Section~\ref{Sec: privacy}. Concepts of fairness and algorithms to ensure it are then discussed in Section~\ref{Sec: fairness}. The intersection of privacy and fairness is the central focus of Section~\ref{Sec: privacy vs fairness}. Open issues and potential directions are explored in Section~\ref{Sec: Vision and Challenges}. Conclusions are drawn in Section~\ref{sec: conclusion}. A comprehensive map, illustrating the sections and subsections, can be found in Appendix A. To the best of our knowledge, this is the first survey that attempts to provide a critical review of privacy and fairness in ML. Some of the most relevant and recent surveys are reviewed in Table~\ref{Table:Related_Works}.

\renewcommand{\arraystretch}{1}


\begin{table}
    \centering
    \caption{Comparison of recent related works and our paper’s key contributions on privacy and fairness in ML.}
    \label{Table:Related_Works}
    \renewcommand{\arraystretch}{1.5}
    \begin{tabular}{|c|c|c|c|c|c|c|c|c|}
    \hline
    \multirow{2}{*}{\textbf{Paper}} 
    & \multirow{2}{*}{\textbf{Key topic}} 
    & \multicolumn{4}{c|}{\textbf{ML categories}} 
    & \multicolumn{2}{c|}{\textbf{Key contributions}} 
    & \multirow{2}{*}{\makecell[c]{\textbf{Highlights}}} \\
    \cline{3-8} & & \makecell{\textbf{SL}} & \makecell{\textbf{UL}} & \makecell{\textbf{SSL}} & \makecell{\textbf{RL}} & \textbf{Privacy} & \textbf{Fairness} & \\
    \hline
    \cite{soykan2022survey} & \makecell[c]{Privacy-preserv-\\ing collaboration\\ in ML} & $\checkmark$ & $\checkmark$ & $\times$ & $\times$ & $\checkmark$ & $\times$ & \makecell[l]{Pioneering study concentrat-\\ing on collaborative ML pri-\\vacy needs and limitations} \\ [0.45cm]
    \hline
    \cite{blanco2022critical} & \makecell[c]{ Limitations of DP \\in ML applications} & $\checkmark$ & $\times$ & $\times$ & $\times$ & $\checkmark$ & $\times$ & \makecell[l]{DP assessment: flaws, trade-\\offs, ML implementation} \\ [0.3cm]
    \hline
    \cite{le2022survey} & \makecell[c]{Fairness-aware ML \\in different datasets} & $\checkmark$ & $\checkmark$ & $\times$ & $\times$ & $\times$ & $\checkmark$ & \makecell[l]{Fairness in ML through in-de-\\pth real data analysis} \\ [0.3cm]
    \hline
    \cite{pessach2022review} & \makecell[c]{Algorithmic fairness\\ in ML} & $\checkmark$ & $\times$ & $\times$ & $\checkmark$ & $\times$ & $\checkmark$ & \makecell[l]{Overview of identifying, mea-\\suring, and improving algorit-\\hmic fairness} \\ [0.45cm]
    \hline
    \cite{choudhary2022survey} & \makecell[c]{Fairness in \\graph mining} & $\checkmark$ & $\checkmark$ & $\checkmark$ & $\times$ & $\times$ & $\checkmark$ & \makecell[l]{Fairness in graph algorithms: \\measures, benchmarks, and \\research directions} \\ [0.45cm]
    \hline
    \cite{liu2021machine} & \makecell[c]{Privacy-preserving \\ML} & $\checkmark$ & $\checkmark$& $\checkmark$ & $\times$ & $\checkmark$ & $\times$ & \makecell[l]{Identifying gaps and challen-\\ges in privacy preservation \\for ML} \\ [0.45cm]
    \hline
    \cite{de2021critical} & \makecell[c]{Privacy defense\\ trade-offs\\ in ML evaluation} & $\checkmark$ & $\times$ & $\times$ & $\times$ & $\checkmark$ & $\times$ & \makecell[l]{Balancing privacy and utility \\in ML defense evaluation} \\
    \hline
    \cite{xu2021privacy} & \makecell[c]{Privacy-preserving \\ML} & $\checkmark$ & $\times$ & $\checkmark$ & $\times$ & $\checkmark$ & $\times$ & \makecell[l]{Integration of privacy techniq-\\ues in ML for data-driven app-\\lications} \\ [0.45cm]
    \hline
    \cite{wan2021modeling} & \makecell[c]{In-processing \\fairness mitigation} & $\checkmark$ & $\checkmark$ & $\checkmark$ & $\times$ & $\times$ & $\checkmark$ & \makecell[l]{Categorization of explicit and \\implicit methods in achieving\\ fairness} \\ [0.45cm]
    \hline
    \cite{mehrabi2021survey} & \makecell[c]{Fairness and bias \\in AI systems} & $\times$ & $\times$ & $\checkmark$ & $\times$ & $\times$ & $\checkmark$ & \makecell[l]{Taxonomy of fairness defini-\\tions for mitigating biases in \\AI} \\ [0.45cm]
    \hline
    \cite{chhabra2021overview} & Fair clustering & $\times$ & $\checkmark$ & $\times$ & $\times$ & $\times$ & $\checkmark$ & \makecell[l]{Organized overview with new \\insights and classifications in\\ fair clustering} \\ [0.45cm]
    \hline
    \cite{tanuwidjaja2020privacy} & \makecell[c]{Privacy-Preserving \\DL in MLaaS} & $\times$ & $\checkmark$ & $\checkmark$ & $\times$ & $\checkmark$ & $\times$ & \makecell[l]{Adversarial models, attacks, \\and solutions in privacy-pre-\\serving DL} \\ [0.45cm]
    \hline
    Our paper & \makecell[c]{Privacy-Fairness \\Interrelation in ML} & $\checkmark$ & $\checkmark$ & $\checkmark$ & $\checkmark$ & $\checkmark$ & $\checkmark$ & \makecell[l]{Thorough review of privacy,\\ fairness in ML, examining im-\\pact, architectures, and re-\\search gaps} \\ [0.65cm]
    \hline
\end{tabular}
\end{table}

\section{Privacy}\label{Sec: privacy}
\subsection{Preliminaries} 

ML has transformed various industries, including healthcare \cite{nayyar2021machine}, transportation \cite{tizghadam2019machine}, and finance \cite{dixon2020machine}. The capacity of these algorithms to process vast quantities of data, identify patterns, generate predictions, and deliver precise and efficient recommendations is remarkable. However, the use of personal data in ML models has given rise to significant privacy concerns. A primary concern is the potential misuse of sensitive personal information \cite{ngiam2019big}, such as names, addresses, social security numbers, and medical records. If not adequately safeguarded, such data could lead to identity theft, financial fraud, and other adverse consequences. The issue of data privacy was starkly highlighted in $2018$ when Cambridge Analytica illicitly harvested data from millions of Facebook users \cite{hinds2020wouldn}. This data was subsequently used to craft effective political advertisements during the 2016 US presidential election~\cite{berghel2018malice}, sparking widespread concern over the use of personal information in political campaigns.

To address these privacy concerns, researchers are developing privacy-preserving algorithms that secure sensitive data while allowing accurate and efficient model creation. Differential privacy (DP) is one such technique, which adds noise to the data to prevent individual records from being identified~\cite{dwork2008differential}. Another technology that allows data to be processed without being decrypted is homomorphic encryption~\cite{acar2018survey}, which ensures that sensitive information is never divulged. While some research have focused on specific privacy concerns connected to certain types of ML learning methodologies, such as supervised, unsupervised, semi-supervised, and RL, a comprehensive evaluation of all potential privacy hazards and remedies is still lacking. Furthermore, there may be privacy issues that are specific to specific domains or applications that necessitate a more concentrated analysis. Consequently, it is necessary to continue exploring privacy concerns in these ML techniques to guarantee that all potential dangers are properly detected and addressed. This will ensure that ML technologies are developed and deployed in a way that respects individuals' privacy rights while also minimizing the possible damages associated with these technologies.

\subsection{Privacy Techniques}
\subsubsection{Differential Privacy (DP)}

DP is a privacy protection method commonly used in different stages of the ML pipeline to enhance privacy of individuals. In this section, we will examine the concepts and definitions, common DP mechanisms, and applications of DP in various ML techniques.
\paragraph{Notions and Definitions}
\begin{defn}[$\epsilon$-Differential Privacy\cite{dwork2008differential}] 
A randomized algorithm $\mathcal{M}$ is said to be $(\epsilon, \delta)$-differentially private if, for any two datasets $D_1$ and $D_2$ that differ in only one data point, and any subset of the range of $\mathcal{M}$, the following holds:
    \begin{equation}
        Pr[M(D_1) \in S] \leq e^{\epsilon} Pr[M(D_2) \in S] + \delta,
    \end{equation}
where $\epsilon$ and $\delta$ are privacy parameters, and $S$ is any subset of the range of $\mathcal{M}$. This inequality ensures that the probability of observing a certain output of $\mathcal{M}$ on a dataset $D_1$ is almost the same as the probability of observing the same output on a dataset $D_2$ that differs in only one data point, with the exception of a small amount of random noise controlled by $\epsilon$ and $\delta$. The parameter $\epsilon$ controls the strength of the privacy guarantee, with lower values providing stronger privacy protection, while $\delta$ is a parameter that accounts for the probability that the privacy guarantee is violated due to the randomness introduced by the algorithm.
\end{defn}

\begin{defn}[$L1$-Sensitivity\cite{dwork2006calibrating}] 
L1-sensitivity is a measure of how much the output of a function changes when a single data point is added or removed from a dataset. It is defined as the maximum absolute difference between the output of the function on two adjacent datasets that differ in only one data point. Formally, given a function $f: \mathcal{D} \rightarrow \mathbb{R}^n$ that maps datasets in domain $\mathcal{D}$ to vectors in $\mathbb{R}^n$, the L1-sensitivity of $f$ is defined as:
\begin{equation}
    \Delta f = \max_{d\in D, d' \sim d} \left\vert\vert f(d) - f(d') \right\vert\vert_1,
\end{equation}
where $d'$ is the neighboring dataset that differs from $d$ by a single data point, and $||\cdot||_1$ denotes the L1-norm. Intuitively, L1-sensitivity captures the largest change that can occur in the output of $f$ due to the presence or absence of a single data point. It is a fundamental parameter in DP, as it determines the amount of noise that needs to be added to the output of $f$ to achieve a desired level of privacy protection.
\end{defn}

\paragraph{Common Mechanisms in DP}
DP approaches involve the addition of controlled noise to data to safeguard the privacy of people while preserving the accuracy of analytic results. The \textit{Laplace mechanism} and \textit{Exponential mechanism} are two often used differentially private mechanisms, which are detailed as follows.
\begin{table*}
    \centering
    \caption{Comparison of Laplace and Exponential Mechanisms in DP.}
    \label{Table:ComparisonDPMechanisms}
    \renewcommand{\arraystretch}{1.4}
    \begin{tabular}{|c|c|c|c|}
        \hline
        \textbf{Mechanism} & \textbf{Description} & \textbf{Pros} & \textbf{Cons} \\
        \hline
        Laplace & \makecell[l]{Adds independent noise drawn from \\a Laplace distribution to the true out-\\put, proportional to the sensitivity of\\ the query and inversely proportional \\to the privacy budget.} & \makecell[l]{Simple implementation, \\provides strong privacy\\ guarantees, and works\\ efficiently for simple qu-\\eries.} & \makecell[l]{Produces noisy results, calib-\\ration of noise parameter can \\be difficult, may not perform\\ well for high-dimensional da-\\ta or complex queries.} \\ 
        \hline
        Exponential & \makecell[l]{Adds independent noise drawn from\\ an exponential distribution to the tr-\\ue output, proportional to the sensiti-\\vity of the query and inversely prop-\\ortional to the privacy budget.} & \makecell[l]{More precise results th-\\an Laplace, can perform\\ well for high-dimensio-\\nal data or complex que-\\ries.} & \makecell[l]{Requires more sophisticated\\ implementation, may be vul-\\nerable to adaptive attacks, \\calibration of noise parame-\\ter can be challenging.} \\ 
        \hline
    \end{tabular}
\end{table*}
\subparagraph{Laplace Mechanism}
The Laplace mechanism \cite{holohan2018bounded} is a method for achieving DP by adding random noise to the output of a query in a way that satisfies DP guarantees. Specifically, given a function $f: D \rightarrow R$ that we want to compute on a dataset $D$, the Laplace mechanism adds random noise to $f(D)$ according to the following formula:
\begin{equation}
    f(D)+Lap\Bigg(\frac{\Delta f}{\epsilon}\Bigg),
\end{equation}
where $Lap(\Delta f/\epsilon)$ is a random variable drawn from the Laplace distribution with mean 0 and scale parameter $\Delta f/\epsilon$, where $\Delta f$ is the sensitivity of the function $f$ and $\epsilon$ is the privacy parameter that controls the amount of noise added. More formally, the Laplace distribution is defined as:
\begin{equation}
    Lap(x \mid \mu, b) = \frac{1}{2b} \exp \left( -\frac{|x-\mu|}{b} \right),
\end{equation}
where $\mu$ is the mean and $b$ is the scale parameter. In the case of the Laplace mechanism, the mean is 0 and the scale parameter is $\Delta f/\epsilon$, so the Laplace distribution becomes:
\begin{equation}
    Lap(x\mid 0,\Delta f/\epsilon)=\frac{1}{2(\Delta f/\epsilon)}\exp\Bigg(-\frac{|x|}{\Delta f /\epsilon}\Bigg).
\end{equation}
Adding Laplace noise to the output of $f(D)$ in this way ensures that the output is differentially private with parameter $\epsilon$. The amount of noise added is proportional to the sensitivity of the function $f$, with higher sensitivities resulting in more noise being added to the output. \newline
Although the Laplace mechanism is commonly used to achieve DP in ML, it has several limitations \cite{koufogiannis2015optimality} \cite{fernandes2021laplace}. The amount of noise added to the data depends on the sensitivity of the function being computed, which can be significant for some functions. This can lead to a considerable loss of output accuracy, making it difficult to obtain meaningful results. Moreover, the Laplace mechanism assumes that the data is continuous and unbounded, which may not always hold for all datasets. Furthermore, the Laplace distribution employed in the mechanism may not be optimal, as it assumes that the noise added to the data is symmetric, which may not be the case in reality. However, this technique is also commonly employed in terms of DP. For example, methods have been devised in \cite{rastogi2010differentially} and \cite{shaham2022differentially} for releasing counts on specific types of data, such as time series. The authors from \cite{xu2013differentially}, \cite{shaham2022htf} and \cite{shaham2021htf} concentrate on releasing histograms, while other authors in \cite{xiao2010differential}, \cite{cormode2012differentially} present ways for reducing the worst-case error of a specified set of count queries.

\subparagraph{Exponential Mechanism}
The exponential mechanism is proposed in \cite{mcsherry2007mechanism}, which is a privacy-preserving approach that selects an item from a dataset based on a specific objective function. It ensures the confidentiality of individuals in the dataset while maximizing the objective function. Formally, let $f : D \to R$ be a function that maps a dataset $D$ to a real number. The exponential mechanism selects an output $d \in D$ with probability proportional to the exponential of the privacy loss incurred by releasing $f(d)$, scaled by a parameter $\epsilon > 0$, which controls the amount of privacy protection:
\begin{equation}
    P(M(D)=d) \propto \exp\left(\frac{\epsilon f(d)}{2\Delta f}\right),
\end{equation}
where $\epsilon$ is the privacy budget and $\propto$ denotes proportionality. The denominator $2\Delta f$ is used to scale the noise so that it is proportional to the sensitivity of the objective function. \newline
In practice, the exponential mechanism is used when we want to select an element from a dataset that satisfies a certain property while minimizing the disclosure of information about the other elements in the dataset. For example, we might want to select a movie from a database that satisfies certain genre preferences while minimizing the disclosure of information about the users who rated the other movies in the database. \newline
Popular way for creating DP in ML, the exponential mechanism has certain limitations\cite{dong2020optimal}. When the dataset is huge, the exponential process can be computationally costly. In addition, as the privacy parameter falls, the precision of the result degrades. In addition, the exponential technique is only appropriate for functions with a low sensitivity value. If the function has a high sensitivity value, the mechanism will add an excessive amount of noise to the output, reducing its precision. In certain instances, such as when the output space is discrete or the objective function is non-convex, the exponential mechanism might be biased. Despite these limitations, this permits DP solutions for a variety of intriguing issues with non-real outputs. As an illustration, the exponential mechanism has been used in the publication of audition results \cite{mcsherry2007mechanism}, coresets \cite{feldman2009private}, support vector machines \cite{rubinstein2009learning}, and frequent patterns \cite{bhaskar2010discovering}.

\paragraph{Spectrum of DP Variations}
\subparagraph{Differential Privacy Stochastic Gradient Descent (DP-SGD)}
DP-SGD emerged from the convergence of two important concepts in the field of ML: Stochastic Gradient Descent (SGD) and DP. SGD is an iterative method for optimizing an objective function and has been extensively used in ML, especially in the training of large-scale deep neural networks. The idea was to provide formal privacy guarantees when disclosing statistical information about a dataset. In 2016, Abadi et al. \cite{abadi2016deep} successfully combined these concepts to develop DP-SGD, a variant of SGD that offers strong privacy guarantees by incorporating differential privacy into the optimization process. 

The DP-SGD algorithm begins by sampling a minibatch from the dataset. For each instance in the minibatch, the gradient $\nabla L(\theta; x)$ of the loss function $L$ with respect to the model parameters $\theta$ is computed. This results in a vector of gradients for the minibatch. The next crucial step in DP-SGD is gradient clipping. This process involves limiting the $L_2$ norm of each individual gradient vector to a predefined threshold $C$. In mathematical terms, this operation can be expressed as:
\begin{equation}
    \nabla L_{\text{{clipped}}}(\theta; x) = \min\left(1, \frac{C}{\|\nabla L(\theta; x)\|}\right) \nabla L(\theta; x).
\end{equation}
This gradient clipping operation ensures that the contribution of each individual instance to the gradient computation is limited, thereby mitigating the impact of outliers and reducing the sensitivity of the output to changes in the input data, a key requirement for DP. After gradient clipping, the algorithm computes the average of the clipped gradients and adds calibrated Gaussian noise to this average. If $G$ represents the average of the clipped gradients, the noisy gradient $G_{\text{{noisy}}}$ is given by:
\begin{equation}
G_{\text{{noisy}}} = G + \mathcal{N}(0, (\sigma C)^2\mathbf{I}),
\end{equation}
where $\mathcal{N}(0, (\sigma C)^2\mathbf{I})$ represents multivariate Gaussian noise with mean 0 and covariance matrix $(\sigma C)^2\mathbf{I}$, and $\mathbf{I}$ is the identity matrix. The model parameters $\theta$ are then updated using this noisy gradient.

DP-SGD is primarily used in scenarios where models need to be trained on sensitive data while preserving privacy. For instance, in healthcare, DP-SGD could be used to build predictive models using patient data without compromising individual privacy \cite{suriyakumar2021challenges}. DP-SGD has also been used in federated learning \cite{geyer2017differentially}, a paradigm where the model is trained across multiple decentralized edge devices, maintaining data on the original device. Several libraries and frameworks have been developed for implementing DP-SGD. Google's TensorFlow Privacy library provides a version of DP-SGD that can be used with TensorFlow models \cite{mcmahan2018general}. Another library is PyTorch-DP (now Opacus) \cite{yousefpour2021opacus}, which provides an implementation for PyTorch models. These libraries provide convenient tools to add privacy-preserving capabilities to ML models with minimal code changes.

\subparagraph{Differential Privacy for Support Vector Data Description (DP-SVDD)}
Support Vector Data Description (SVDD) \cite{tax2004support} is a one-class classification method that is often used for anomaly detection. The main idea behind SVDD is to find a hypersphere in the feature space that encapsulates the majority of the data points. This hypersphere is described by its center and radius, and it is found by solving an optimization problem that aims to minimize the radius while penalizing data points that lie outside the hypersphere. Mathematically, the SVDD problem can be formulated as follows:
\begin{align}
\text{Minimize:} \quad & R^2 + C \sum \xi_i, \\
\text{Subject to:} \quad & ||\phi(x_i) - a||^2 \leq R^2 + \xi_i \quad \text{and} \quad \xi_i \geq 0.
\end{align}
In this formulation, $R$ is the radius of the hypersphere, $a$ is the center, $\phi(x_i)$ is the mapping of data point $x_i$ in the feature space, $\xi_i$ is the slack variable that allows data points to lie outside the hypersphere, and $C$ is a regularization parameter that controls the trade-off between the volume of the hypersphere and the errors.
DP-SVDD, first introduced in \cite{park2023efficient}, is a method that combines the principles of SVDD with those of DP to create a privacy-preserving one-class classification model. The method involves two main phases. In the first phase, the goal is to train a SVDD model while ensuring differential privacy. The center of the hypersphere in the SVDD model is represented as a weighted sum of the mapped data points. To ensure DP, the center of the hypersphere is perturbed by adding noise. This noise is drawn from a Laplace distribution. The perturbed center, $\hat{a}$, is then given by:
\begin{equation}
\hat{a} = a + l = \sum_{i=1}^{n} b_i \phi(x_i) + l.
\end{equation}
In this formulation, $a$ is the center of the hypersphere, $b_i$ are the dual variables, $\phi(x_i)$ is the mapping of data point $x_i$ in the feature space, $l$ is the Laplace noise. The sensitivity is a measure of how much the output of a function can change when a single data point is added or removed from the dataset. In the second phase, the input space is partitioned into separate regions using a dynamical system based on the differentially private support function from the first phase. This dynamical system is defined by the gradient of the support function:
\begin{equation}
\frac{dx}{dt} = \nabla \hat{f}(x),
\end{equation}
where $\hat{f}(x)$ is the differentially private support function. Regions, associated with Equilibrium Points (EPs) of the dynamical system, are labeled using a noisy count of class labels of converging data points. The privacy-preserving predictions are released by publishing private EPs and labels. A new data point's label is predicted based on its region, determined by the EP it converges to. The privacy of predictions is ensured by the differential privacy of the support function and the noisy count.


The preceding discussion regarding the DP has improved our understanding of the definitions and mechanisms of the DP, as well as the limitations of each mechanism, which are summarized in Table \ref{Table:ComparisonDPMechanisms}. Additionally, a comprehensive examination of the real-world applications of DP in supervised, unsupervised, semi-supervised, and reinforcement learning is presented in Appendix B.

\subsubsection{Homomorphic Encryption (HE)}
Homomorphic Encryption (HE) represents a cryptographic technique that confers the capacity to perform computations on encrypted data without the need for decryption. Such an approach stands out as a promising means for preserving data privacy while enabling useful computations to be conducted on it. The following section reviews and categorizes the concepts, prevalent HE mechanisms, and a range of applications of HE in the context of ML techniques.
\paragraph{Notions and Definitions}
\begin{defn}[Homomorphic Encryption\cite{yi2014homomorphic}] 
Homomorphic encryption enables computations to be performed on ciphertexts without the need to decrypt them first. Mathematically, let $f$ be an algebraic function and $Enc$ and $Dec$ be encryption and decryption functions respectively. Homomorphic encryption allows for the following equation to hold:
    \begin{equation}
f(\operatorname{Dec}_k(\operatorname{Enc}_k(m_1))\circ \operatorname{Dec}_k(\operatorname{Enc}_k(m_2)))=\operatorname{Enc}_k(f(m_1\circ m_2)),
    \end{equation}
In this equation, $m_1$ and $m_2$ are plaintext messages that are encrypted under the same key $k$ using HE. The function $f$ is a homomorphic function that operates on the plaintext messages, and the operator $\circ$ represents the algebraic operation that $f$ preserves. The equation shows that applying $f$ to the plaintext messages $m_1$ and $m_2$ and then encrypting the result under the key $k$ is equivalent to first encrypting the plaintext messages separately, applying the decryption function $\mathrm{Dec}_k$ to each ciphertext, performing the algebraic operation $\circ$ on the resulting plaintexts, and then encrypting the result again under the key $k$. This property allows computations to be performed on encrypted data without ever revealing the plaintext to the party performing the computation.
\end{defn}

\paragraph{Typical Schemes in HE}
There exist various schemes of homomorphic encryption, each possessing its unique merits and demerits. The \textit{Fully Homomorphic Encryption}, \textit{Partially Homomorphic Encryption}, and \textit{Somewhat Homomorphic Encryption} are three frequently utilized HE schemes, explicated as follows.

\begin{table*}
    \centering
    \caption{Comparison of Homomorphic Encryption Schemes.}
    \label{Table:ComparisonHESchemes}
    \renewcommand{\arraystretch}{1.4}
    \begin{tabular}{|c|c|c|c|}
        \hline
        \textbf{Property} & \textbf{FHE} & \textbf{PHE} & \textbf{SHE} \\
        \hline
        Supports addition & $\checkmark$ & $\checkmark$ & $\checkmark$ \\ 
        \hline
        Supports multiplication & $\checkmark$ & $\times$ & $\checkmark$ \\ 
        \hline
        \makecell{Supports arbitrary \\circuits} & $\checkmark$ & $\times$ & $\times$ \\ 
        \hline
        \makecell{Computational \\complexity} & High & Moderate & Low \\ 
        \hline
        \makecell{Encryption/decryption\\ speed} & Slow & Moderate & Fast \\ 
        \hline
        Application examples & \makecell{Cloud computing, privacy-\\preserving machine learning} & \makecell{Secure multi-party compu-\\tation, secure function eva-\\luation} & \makecell{Privacy-preserving data \\analysis, secure computa-\\tion protocols} \\ \hline
    \end{tabular}
\end{table*}

\subparagraph{Fully Homomorphic Encryption (FHE)}
FHE allows computations on encrypted data without decryption, leading to direct computation on ciphertexts and yielding an encrypted plaintext. It employs lattice-based cryptography \cite{micciancio2009lattice} with ideal lattices to efficiently compute homomorphic operations. This type of cryptography is based on mathematical lattices, regular patterns of points. FHE carries out encryption and decryption on ideal lattices, which are sets of linear combinations of $n$ independent vectors with integer coefficients in $n$-dimensional space. Mathematically, this can be expressed as:
\begin{equation}
    L = {a_1.v_1 + a_2.v_2 + ... + a_n.v_n | a_i \in Z},
\end{equation}
where $L$ is the lattice, $v_1$, $v_2$, ..., $v_n$ are linearly independent vectors in $n$-dimensional space, and $a_1$, $a_2$, ..., $a_n$ are integers. The security of FHE is based on the hardness of certain problems related to lattices, such as the Shortest Vector Problem (SVP) and the Closest Vector Problem (CVP) \cite{hanrot2011algorithms}. These problems are known to be difficult to solve in high dimensions, which provides the basis for the security of FHE.
\newline
To perform homomorphic operations on ciphertexts in FHE, a technique called "bootstrapping" or "gating" is used. This technique involves decrypting the ciphertext using the secret key, performing a homomorphic operation on the resulting plaintext, and then encrypting the result using the public key. Mathematically, this can be represented as:
\begin{equation}
    C' = Enc_{pk}(F(Dec_{sk}(C))),
\end{equation}
where $C$ is the original ciphertext, $sk$ is the secret key, $pk$ is the public key, $F$ is the homomorphic operation, $Dec_{sk}(C)$ is the decrypted ciphertext, and $Enc_{pk}$ is the encryption function using the public key.
\newline
FHE presents a spectrum of advantages and disadvantages \cite{yousuf2020systematic}. On the one hand, FHE confers a paramount level of security as it allows for arbitrary computations on encrypted data without necessitating decryption. This characteristic proves exceptionally advantageous in domains such as ML and cloud computing, where privacy and security concerns are of utmost importance. Conversely, FHE exhibits a high level of computational complexity that can render it infeasible for certain applications \cite{ogburn2013homomorphic}. Moreover, with each homomorphic operation, the size of the ciphertext augments, requiring extensive memory, which can become a significant impediment \cite{kogos2017fully}. Despite these challenges, FHE remains an active area of research and development, with researchers continuously seeking ways to enhance its efficiency and transcend its limitations.

\subparagraph{Partially Homomorphic Encryption (PHE)}
Partially Homomorphic Encryption (PHE) is a type of encryption scheme that enables computation on encrypted data without the need to decrypt it \cite{moore2014practical}. Mathematically, PHE is defined using algebraic structures such as groups, rings, or fields to enable certain types of computation, such as addition or multiplication, on ciphertexts while still maintaining the confidentiality of the underlying plaintext \cite{sen2013homomorphic}. PHE schemes can be partially homomorphic, meaning that they support computations of only one type, such as addition or multiplication. For example, a PHE scheme that is homomorphic with respect to addition is defined by the following property:
\begin{equation}
    Enc(m_1) + Enc(m_2) = Enc(m_1 + m_2),
\end{equation}
where $m_1$ and $m_2$ are plaintext messages, $Enc$ is the encryption function, and $+$ denotes addition in the plaintext space $M$. This property allows ciphertexts to be added together and then decrypted to obtain the sum of the corresponding plaintexts. Similarly, a PHE scheme that is partially homomorphic with respect to multiplication is defined by the following property: 
\begin{equation}
    Enc(k \cdot m) = Enc(m)^k,
\end{equation}
where $k$ is a scalar value and $m$ is a plaintext message. This property allows a ciphertext to be raised to a scalar power $k$ without revealing the plaintext, but it does not allow multiplication of two ciphertexts to obtain a ciphertext that represents the multiplication of the corresponding plaintexts.
\newline
In reality, PHE is a powerful cryptographic technique that offers many benefits \cite{yu2012partial}. PHE allows for computations on encrypted data, enabling secure processing of sensitive data without its disclosure to unauthorized parties. Unlike FHE, PHE does not require heavy computational resources and is, therefore, much easier to implement in real-world applications. PHE can be implemented with relatively straightforward mathematical operations, making it both basic and effective. In addition, PHE can be used to build secure protocols for a range of applications, such as secure auctions, electronic voting, and secure multi-party computation. By keeping the data encrypted, PHE can ensure that sensitive information remains private while allowing authorized parties to perform meaningful computations on it. While there are some limitations to PHE \cite{hellwig2022distributed}, such as its limited computational capacity and susceptibility to attack if not implemented correctly, the benefits of this technique make it a valuable tool in a variety of situations.

\subparagraph{Somewhat Homomorphic Encryption (SHE)}
Somewhat Homomorphic Encryption (SHE) is a form of encryption that permits certain calculations on encrypted data without revealing the original data \cite{fan2012somewhat}. Using a polynomial representation of the plaintext and encrypting it with a public key is a central concept of SHE \cite{boneh2013private}. By manipulating the coefficients of the polynomial, it is possible to perform computations on the ciphertext. A commonly used example of a SHE scheme is the BGV (Bajard, Gentry and Vaikuntanathan) scheme \cite{aguilar2013recent}. Consider the BGV scheme, which operates over the polynomial ring $R_q = \mathbb{Z}[x]/(xn + 1)$, where $q$ is a prime number that determines the security level and $n$ is the degree of the polynomial. $R_2$ denotes the set of polynomials with coefficients in $\{0,1\}$, which is the plaintext space. 
\newline
To encrypt a plaintext polynomial $m(x)$, the BGV scheme first generates a random polynomial $r(x)$ with coefficients in $\{0,1\}$. It then computes the ciphertext polynomial $c(x)$ as:
\begin{equation}
    c(x) = r(x) * pk + m(x) * 2^k \text{ (mod } q \text{)},
\end{equation}
where $pk$ is the public key, $k$ is a positive integer, and * denotes polynomial multiplication. The random polynomial $r(x)$ serves as a noise term that hides the underlying plaintext, while the term $m(x) * 2^k$ ensures that the ciphertext coefficients are sufficiently large to prevent decryption attacks. Conversely, to decrypt a ciphertext $c(x)$, one needs to compute:
\begin{equation}
    m(x) = c(x) * sk \text{ (mod } q \text{)} \text{ mod } 2,
\end{equation}
where $sk$ is the secret key. The term $c(x) * sk$ cancels out the noise term $r(x)$, and yields the original plaintext polynomial $m(x)$. Finally, to perform a computation on two ciphertexts $c_1(x)$ and $c_2(x)$, one can simply add or multiply them as polynomials, and obtain the resulting ciphertext $c_3(x)$ as:
\begin{equation}
    c_3(x) = c_1(x) + c_2(x) \text{  or 
 }c_3(x) = c_1(x) * c_2(x).
\end{equation}
Though SHE can enable computations to be performed on encrypted data without requiring the data to be decrypted, this scheme suffers from certain limitations that affect its practicality in certain scenarios \cite{hamza2022towards}. These limitations stem from its lack of full homomorphic capabilities, which constrain the range of computations that can be performed. Furthermore, SHE is typically associated with higher computational overheads and requires more computational resources, which can affect its overall efficiency and practicality. Nevertheless, SHE offers several benefits \cite{migliore2018practical}, such as preserving the privacy of sensitive data, while allowing computations to be performed in a secure and confidential manner. This makes SHE a useful technique in scenarios where privacy and security are paramount, such as in the healthcare and financial sectors. Additionally, SHE can be used in conjunction with other cryptographic techniques, such as fully homomorphic encryption, to provide a more comprehensive security framework. Hence, despite its limitations, SHE remains a valuable and promising technique in the field of cryptography.


The preceding discourse on HE has enhanced our knowledge of its definitions, mechanisms, and the limitations of each scheme, as detailed in Table \ref{Table:ComparisonHESchemes}. Additionally, a thorough examination of the practical applications of HE in supervised, unsupervised, semi-supervised, and reinforcement learning is provided in Appendix C.\\


\section{Fairness}\label{Sec: fairness}
\subsection{Preliminaries}

The concept of fairness in society has been a recurring study subject throughout history~\cite{anderson2004pursuit}. Although early discussions were mainly philosophical, the rise of data and ML in the past decade has attracted tremendous attention to fairness in algorithms. As opposed to the initial perception of models and algorithms being trustworthy, soon it was realized that they could lead to severe unjust decisions, affecting especially individuals from disadvantaged groups. Perhaps, the most significant of such discoveries was revealed in an article published by ProPublica in $2016$, highlighting the significance of algorithmic fairness. The article focuses on a software named COMPAS~\cite{wenger2011compass} designed to determine the risk of a person committing another crime and assist US judges in making release decisions. The investigation found that COMPAS was biased against African Americans as it had a higher rate of false positives for this group compared to Caucasians. This and numerous other examples indicate the necessity to quantify and mitigate unfairness-related issues in ML. In the remaining of this subsection, we discuss bias in ML, what the law says about the issue, and online tools available to address the problem.

\subsubsection{Bias}

The term "bias" in ML has a distinct meaning that is different from the typical understanding of the term in social and news contexts~\cite{campolo2017ai}. Bias is seen as the root cause of unfairness and is often tied to a specific term that indicates where in the process, the data is being distorted. Over time, many different types of bias have been introduced in the literature, some of which are subcategories of others, leading to confusion in properly defining each one. For interested readers, we have provided a thorough classification and visualization of bias in ML in Appendix D. In this categorization, we have grouped potential types of bias into four general categories: \emph{A Biased World}, \emph{Data Collection and Preparation}, \emph{Model Training}, and finally, \emph{Evaluation and Deployment}.

\subsubsection{Philosophies of Fairness in Context}

In the domain of work and employment, principles of fairness and non-discrimination guide the relationships among employees, employers, and labor unions. Two core fairness principles, often identified as `Disparate Impact' and `Disparate Treatment', are observed in this context. Disparate Treatment~\cite{zimmer1995emerging} acknowledges that unjust behaviors towards individuals due to their protected attributes, such as race, are unacceptable. An instance reflecting this principle in action could be prohibiting the exclusive skill examination of job applicants based on their ethnic group affiliation. Disparate impact~\cite{rutherglen1987disparate} pertains to practices that inadvertently disadvantage a protected group, even though the policies implemented by organizations appear neutral on the surface. This principle recognizes that discrimination is not always direct, and it can affect individuals and groups in indirect ways. A classic example includes policies that, while appearing neutral, disproportionally impact members of a protected group in a negative manner~\cite{selbst2017disparate}.

In ML, fairness principles are implemented in various ways to uphold the aforementioned principles for sensitive attributes. Notably, different organizations provide guidelines on what constitutes sensitive attributes. The most commonly protected features include race, gender, religion, and national origin. For more detailed information, please refer to Appendix E, which provides a table of sensitive attributes identified by several organizations.


\subsubsection{Available Online Tools}

The significance of algorithmic fairness has led to the development of several online tools to assist the incorporation of fair practices in ML models. \emph{Fairlearn} \cite{bird2020fairlearn} is an open-source tool developed at Microsoft Research that helps data scientists and developers assess and improve the fairness of their AI models. \emph{AI Fairness 360} (AIF360) \cite{bellamy2019ai} is an open-source toolkit created by IBM that helps developers analyze, document, and eliminate unfairness in ML models throughout their lifecycles. It provides various metrics and mitigation algorithms to assess and address bias in models. \emph{Aequitas} \cite{saleiro2018aequitas} is another open-source tool that helps identify and eliminate bias in ML models. It offers metrics, visualizations, and techniques for auditing models, allowing researchers, analysts, and policymakers to make informed decisions during model development and deployment. Google's \emph{What-If Tool} \cite{wexler2019if} and LinkedIn Fairness Toolkit (LiFT) \cite{vasudevan2020lift} are also some of the latest developments to further enhance algorithmic fairness.

\textbf{Structure}
In the following subsections, we thoroughly review fairness in supervised, unsupervised, semi-supervised, and RL. In each subsection, we start by defining fairness notions and definitions dedicated to the type of ML learner, followed by explaining the existing unfairness mitigation techniques for fair treatment of individuals and groups. The mitigation algorithms are divided into pre-processing, in-processing, and post-processing strategies. SSL lies at the intersection of supervised and unsupervised learning. To the best of our knowledge, there is no specific fairness notion proposed particularly for SSL, despite the existence of dedicated unfairness mitigation algorithms. Due to limited space, we have moved the discussions related to SSL to Appendix F. 

\subsection{Fairness in Supervised Learning}
\subsubsection{Notions and Definitions}

As opposed to privacy, where at least for statistical databases, there is a consensus on DP, there does not exist such an agreement on a common notion for fairness. One suggested guideline is to select the notion based on the underlying application. This section reviews some of the most widely adopted fairness notions for supervised learning. In this context, we use terms notion and definition interchangeably. Also, deviation from a fairness notion is referred to as {\em discrimination level}. Discrimination is usually manifested as the absolute value of the difference in metrics for different groups. Moreover, we denote the set of sensitive attributes by $A$, all observed attributes by $X$, latent attributes not observed by $U$, true label to be predicted by $Y$, and finally, predictor by $\hat{Y}$.

We debut our discussions on notions with statistical parity, one of the primary group-level fairness notions. 

\begin{defn} \label{def: statistical parity}
(Demographic or Statistical Parity~\cite{dwork2012fairness,calders2010three}).  A predictor $\hat{Y}$ satisfies demographic parity if:
\begin{equation}
    P(\hat{Y}=1| A = 0) = P(\hat{Y}=1|A=1).
\end{equation}
\end{defn}

\begin{figure}[t]
\includegraphics[scale=.2]{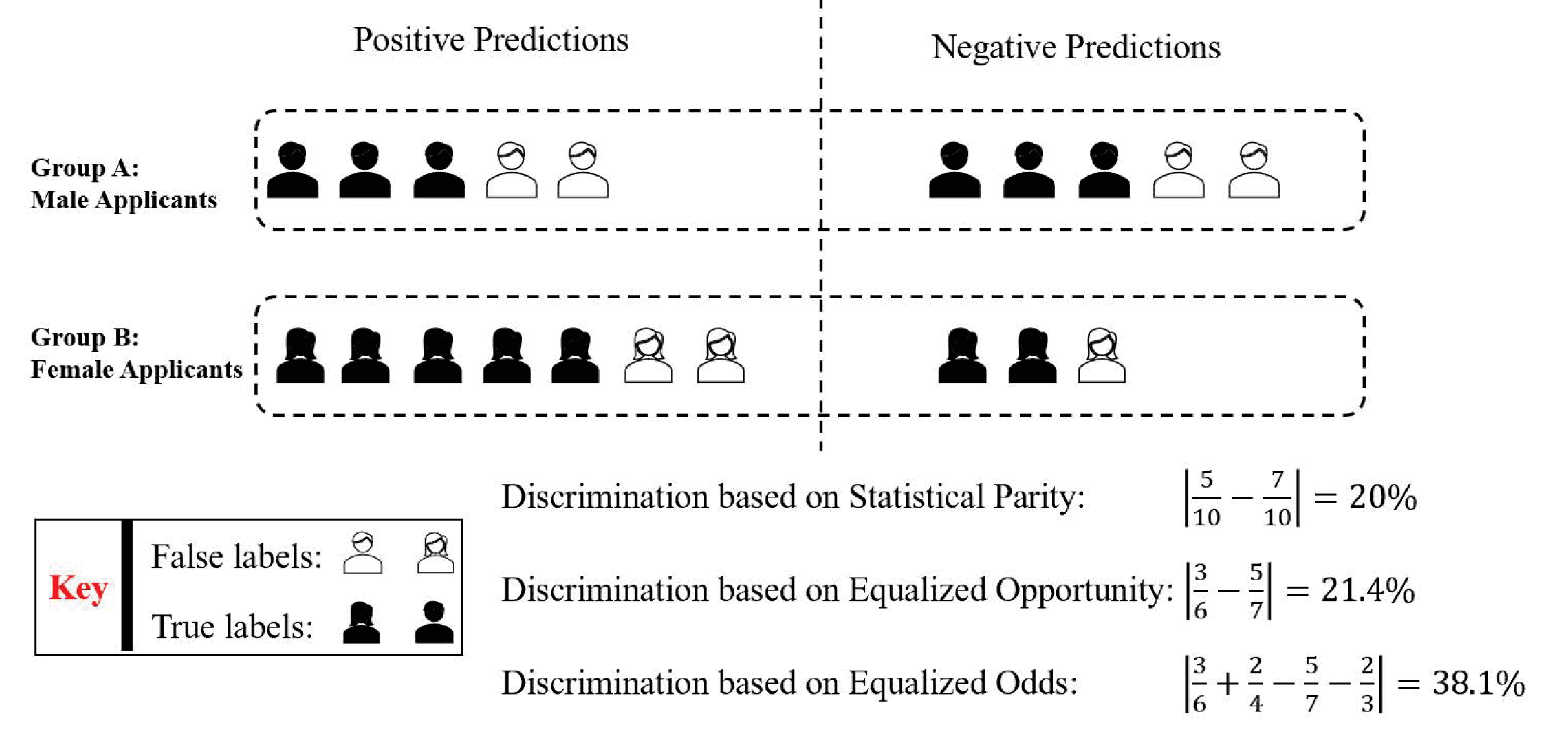}
\centering
\caption{Computation of discrimination based on different metrics.}
\vspace{-15pt}
\label{fig: statistical parity example}
\end{figure}

Statistical parity dictates that regardless of an individual's group, they should have an equal chance of being assigned to a positive class. Figure~\ref{fig: statistical parity example} exemplifies statistical parity. Consider two groups of male and female job applicants and an ML model that decides whether a person should proceed for further evaluation in their application. Here, the likelihood of moving ahead with male and female applicants is $5/10$ and $7/10$, respectively. Hence, discrimination based on statistical parity is $20\%$. The notion of equalized odds, presented next, takes a step further and requires an equal true positive rate across groups.

\begin{defn} \label{def: Equalized Opportunity}
(Equalized Opportunity~\cite{dwork2012fairness}). A predictor $\hat{Y}$ satisfies equal opportunity with respect to protected attribute $A$ and outcome $Y$, if $\hat{Y}$ and $A$ are independent conditional on $Y$,
\begin{equation}
    P( \hat{Y}=1|A=0,Y =1) = P( \hat{Y}=1|A=1,Y =1).
\end{equation}
\end{defn}

That means the true positive rate should be the same for both groups. Going back to the example in Figure~\ref{fig: statistical parity example}, the true positive rate for males and females is $3/6$ and $5/7$, leading to discrimination of $21.4\%$ based on equalized opportunity. The next notion, equalized odds, dictates an even stricter fairness notion requiring equal true and false positive rates across groups.

\begin{defn} \label{def: Equalized Odds}
(Equalized Odds~\cite{dwork2012fairness}). A predictor $\hat{Y}$ satisfies equalized odds with respect to protected attribute $A$ and outcome $Y$~if:
\begin{equation}
    P( \hat{Y}=1|A=0,Y =y) = P( \hat{Y}=1|A=1,Y =y),\;\; y\in \{0,1\}.
\end{equation}
\end{defn}

In the example, discrimination based on equalized odds is $38.1\%$. As can be seen, imposing higher fairness guarantees intuitively results in a higher percentage of discrimination. 


Definition~\ref{def: Calibration} introduces the concept of calibration, which is a crucial idea borrowed from ML. This notion ensures that the confidence scores produced by the model can be interpreted as probabilities and is considered a group-level fairness notion.

\begin{defn} \label{def: Calibration}
(Calibration~\cite{flores2016false,pleiss2017fairness}).  An ML model is said to be calibrated if it produces calibrated confidence scores. Formally, the outcome score $R$ is said to be calibrated if for all the scores $r$ in the support of $R$ following stands, 
\begin{equation}
    P( y = 1 | R = r  )= r.
\end{equation}
\end{defn}

Calibration ensures that the set of all instances assigned a score value $r$ has an $r$ fraction of positive instances among them. Note that the metric is defined on a group level, and it does not mean that an individual who has a score of $r$ corresponds to $r$ probability of a positive outcome. For example, given $10$ people who are assigned a confidence score of $0.7$, in a well-calibrated model, we expect to have $7$ individuals with positive labels among them. 

So far, the fairness definitions discussed were all focused on group-level fairness. In the following, two of the common notions to achieve fairness at an individual level are presented.

\begin{defn} \label{def: counterfactual fairness}
(Counterfactual Fairness~\cite{kusner2017counterfactual}). Given a causal model ($U$, $V$, $F$), where $U$, $V$, and $F$ represent the set of latent (unobserved) background variables, the set of observable variables, and a set of functions defining the mapping $U \cup V \rightarrow V$, respectively, a predictor $\hat{Y}$ is considered counterfactually fair if, under any context $X = x$ and $A = a$, the following equation holds:

\begin{equation}
P(\hat{Y}_{A\leftarrow{a}} (U)=y | X=x, A=a) = P(\hat{Y}_{A\leftarrow{a'}} (U)=y | X=x, A=a).
\end{equation}

This holds for all $y$ and for any value $a'$ attainable by $A$. Here, $A$, $X$, and $\hat{Y}$ represent the set of sensitive attributes, remaining attributes, and decision output, respectively. In other words, the model's predictions for a person should not change in a counterfactual world in which the person's sensitive features are different.
\end{defn}

\begin{defn} \label{def: Individual location fairness}
(Individual Fairness by Dwork et al.~\cite{dwork2012fairness}). 
For a mechanism $\mathcal{M}$ mapping $\boldsymbol{u}$ in the input space $\mathcal{X}$ to value $y$ in the output space $\mathcal{Y}$, individual fairness is satisfied when for any $\boldsymbol{u},\boldsymbol{v} \in \mathcal{X}$:
\begin{equation}
     d_{\mathcal{X}}(\boldsymbol{u},\boldsymbol{v})\geq d_{\mathcal{Y}}(\mathcal{M}(\boldsymbol{u}),\mathcal{M}(\boldsymbol{v})),
\end{equation}
where $d_{\mathcal{X}}: \mathcal{X} \times \mathcal{X} \rightarrow \mathbb{R}_+$ and $d_{\mathcal{Y}}: \mathcal{Y} \times \mathcal{Y} \rightarrow \mathbb{R}_+$.
\end{defn}

To illustrate, let's consider the scenario of classification, where the classifier's predictor $\hat{Y}$ serves as the mapping mechanism. In the context of individual fairness, the fundamental idea is that two individuals who are alike in relevant ways should receive comparable outcomes. To operationalize this concept, we rely on two crucial distance metrics: (1) a similarity distance metric $d_{\mathcal{X}}$ that gauges how similar two individuals are to each other, and (2) a distance metric $d_{\mathcal{Y}}$ that quantifies the disparity between the distributions of outcomes.

\subsubsection{Unfairness Mitigation Algorithms}

\paragraph{Pre-processing Strategies}

{\bf Reweighting.} This approach focuses on modifying the significance of data points manifested as weights during the training to mitigate bias. The method employed in \cite{kamiran2012data} estimates the probability of an individual from a group receiving a specific result and employs the ratios of these probabilities as reweighting factors during optimization. Nevertheless, in some instances, access to sensitive information may be restricted. To tackle this issue, Lahoti et al.~\cite{lahoti2020fairness} learn the reweighting factor using adversarial learning. Furthermore, Roh et al. \cite{roh2020fairbatch} suggest a two-tier optimization method that chooses specific mini-batch sizes to attain group fairness.

{\bf Representation learning.} More recently, a thread of research has focused on changing data representation to improve fairness in classifiers. Several techniques have been proposed predicated on either philosophy of fairness through unawareness or fairness through awareness. In the former, the logic is to learn a new representation for individuals independent of their protected attribute while preserving other information, and the latter focuses on achieving fairness while considering protected group information. In~\cite{feldman2015certifying}, the authors modified dataset features to have similar distributions for both protected and unprotected groups, making it hard to distinguish between them. Zemel et al.~\cite{zemel2013learning} proposed mapping individuals to a new distribution that protects the protected group information while retaining other information. The authors in~\cite{louizos2015variational} follow up this idea by using a variational auto-encoder to make the sensitive attributes independent of the latent representation and applying further learning to that representation. 

{\bf Label alteration.} Label alteration aims to make classifiers fairer by adjusting the labels of training samples. Some works modify the labels to achieve an equal proportion of positive examples across protected groups \cite{kamiran2012data}, while others flip the labels of instances that have been determined to be discriminatory based on differences in treatment among similar samples \cite{luong2011k}.


\paragraph{In-processing Strategies}

{\bf Regularizers and constraints.} This strategy aims to add penalty terms to the classifier's objective function to either minimize the impact of sensitive features on prediction~\cite{kamishima2012fairness}, or achieve similar False Positive and False Negative rates across populations~\cite{bechavod2017penalizing, berk2017convex}. In a similar approach, Kamiran et al.~\cite{kamiran2010discrimination} propose modifying the splitting criterion in decision trees to minimize the impact of sensitive features and maximize the information gain between the split feature and class label. The authors in several studies aim to enforce fairness notions constraints in optimization. This approach for statistical parity is discussed in~\cite{zafar2017fairness}, equalized odds and opportunity in~\cite{zafar2017fairness2, woodworth2017learning}. Quadrianto et al.~\cite{quadrianto2017recycling} suggest using privileged learning to ensure fairness where sensitive features are only available during training. 

{\bf Adversarial learning.} The main concept within this group involves utilizing Generative Adversarial Networks (GANs) to optimize the effectiveness of a predictor while reducing its capability to forecast sensitive characteristics~\cite{zhang2018mitigating}. This approach can be implemented across various gradient-based learning models, such as classification and regression assignments.

{\bf Reweighting.} The reweighting approach proposed part of pre-processing strategies has also been employed during the training. The proposed approach by Krasanakis et al. \cite{krasanakis2018adaptive} trains an unweighted classifier, then learns weights for each sample and retrains the classifier to improve the fairness-accuracy trade-off. The iterative approach of improving reweighting factors helps to derive more accurate reweighting factors.

\paragraph{Post-processing Strategies} 

{\bf Transformation.} This technique aims to modify the output scores to achieve higher fairness levels. One of the primary techniques in this category is Platt Scaling focused on improving miscalibration in ML models~\cite{platt1999probabilistic}. Calibration is improved by fitting the output scores to a logistic regression model. Histogram Binning and Isotonic Regression~\cite{zadrozny2001obtaining} also enhance calibration by fitting output scores to a monotonic function. To achieve individual fairness, the authors in~\cite{shaham2022models} propose using $c$-fair polynomials, which map classifier scores to a polynomial and restrict scores of each individual by their sensitive feature distance. Petersen et al.~\cite{petersen2021post} improve individual fairness by smoothing output scores using a similarity graph and Laplacian regularization. Kim et al.~\cite{kim2019multiaccuracy} present a Multi-accuracy Boost framework that improves accuracy across all subgroups, using iterations and weights to enhance predictions conducted by the auditor.

{\bf Thresholding.} The proposed techniques in this category aim to adjust the label generation threshold of classifiers to make non-discriminatory decisions \cite{kamiran2012decision}. For instance, different threshold values are selected for protected groups in \cite{menon2017cost} to maximize accuracy while achieving statistical parity. Hardt et al.~\cite{hardt2016equality} optimizes threshold selection for each sensitive group for high utility and improved fairness. Similarly, the authors in~\cite{corbett2017algorithmic} infer group-specific thresholds for a trade-off between accuracy and fairness. Lohia et al.~\cite{lohia2019bias} develop a bias-mitigation technique by targeting samples that inhibit individual bias from improving individual and group-level fairness notions.

\subsection{Fairness in Unsupervised Learning}

When data labels are unavailable, unsupervised ML algorithms are commonly used as opposed to supervised algorithms. However, evaluating fairness is more challenging in the absence of labels because there is no ground truth available for assessment. To address this issue, we will begin our discussion by examining individual and group-level fairness notions that have been suggested for unsupervised learning, followed by an exploration of mitigation algorithms.

\subsubsection{Notions and Definitions}

The majority of fairness concepts for unsupervised learning are based on the disparate impact doctrine, which seeks to achieve a comparable proportion of protected groups across all clusters. To begin our conversations on fairness concepts, we will start by examining the Balance metric, regarded as one of the fundamental definitions of group-level fairness in unsupervised learning.

\begin{defn}
(Balance\cite{chierichetti2018fair,bera2019fair}). Define the ratio of the protected group $b\in [m]$ in the entire dataset as $r_b$, and let $r_{a,b}$ represent this proportion in the generated cluster $a\in [k]$. The balance metric evaluates the disparity between these two ratios by defining $R_{a,b} = r_b/r_{a,b}$ and introducing the balance fairness concept as follows:

\begin{equation}
\min_{a\in [k], b\in [m]} \min { R_{a,b}, 1/R_{a,b} }.
\end{equation}
\end{defn}

The Balance metric produces values within the range of $0$ to $1$, where higher scores indicate a higher level of fairness. This measure takes into account both the percentage of protected group members in the entire dataset and within individual clusters, with fairness achieved when the ratio remains consistent across all clusters.

\begin{defn}
(Bounded Representation\cite{ahmadian2019clustering}). Let $r_{a,b}$ denote the ratio of protected group $b\in [m]$ in cluster $a\in [k]$. The $(\alpha,\beta)$-bounded representations dictates that:

\begin{equation}
    \beta \leq   r_{a,b}   \leq \alpha.
\end{equation}

\end{defn}

Bounded representation allows some degree of deviation in the proportion of protected groups within clusters. When the bounds are equal, it suggests that the proportion of protected groups in each cluster should be consistent with the overall ratio in the dataset.

\begin{defn} \label{def: MFC}
(Max Fairness Cost (MFC)~\cite{chhabra2020fair}). Let $I_b$ denote the ideal proportion of protected group $b\in [m]$ in clusters. Once the ideal ratio parameter is passed as input, the MFC notion is defined as 

\begin{equation}
    \max_{a\in [k]} \sum_{b\in[m]} |r_{a,b}-I_b|,
\end{equation}
where $r_{a,b}$ denotes the ratio of protected group $b$ in cluster $a\in [k]$. 
\end{defn}

Intuitively, MFC calculates the summation of all deviations from the ideal ratios for each protected group, and returns the maximum value. A lower MFC value indicates a higher degree of fairness. Setting the value of $I_b$ equal to the ratio of the protected group in the original dataset ($r_b$) ensures that this ratio remains consistent across all clusters.

\begin{defn} \label{def: Social Fairness}
(Social Fairness~\cite{ghadiri2021socially}). Let $C$ denote the cluster centers in $k$-means algorithm and $L(C,D_b)$ denote the $k$-means clustering cost, where $D_b$ is the input error on the samples of the protected group $b\in [m]$. The social fairness notion is then defined as:
\begin{equation}
    \max_{b\in [m]} \dfrac{L(C,D_b)}{|D_b|},
\end{equation}
\end{defn}
Social fairness focuses on the maximum imposed loss on protected groups. Next, we focus on fairness notions proposed on the individual-level.

\bigskip

\bigskip

\begin{defn} \label{def: Fuzzy Individual Fairness}
(Fuzzy Individual Fairness~\cite{dwork2012fairness}). For every two points $x$ and $y$, and their respective distributions $X$ and $Y$ over clusters in a given fuzzy clustering algorithm, let $F(x,y)$ measure the similarity between the two datapoints, and let $D_f(X||Y)$ denote the statistical distance between their distributions. Fuzzy individual fairness requires the satisfaction of the following constraint:

\begin{equation}
D_f(X||Y) \leq F(x,y).
\end{equation}
\end{defn} 

This notion aims to apply the individual fairness notion in~\cite{dwork2012fairness} for fuzzy clustering. Common choices for measuring the statistical distance include the variations of $f$-divergence metric, such as KL-divergence, reverse KL-divergence, and the total variation distance.

\bigskip

\begin{defn} \label{def: kleindessner2020notion}
(Individual Fairness~\cite{kleindessner2020notion}). This notion requires the average distance of every sample point to members in its own cluster to be smaller than its average distance to members of any other cluster. Formally, for a disjoint clustering of data denoted by $\mathcal{C} = { C_1,, C_2,, \ldots, C_k }$, and a distance metric $d$, for every sample point $x\in C_i$, the following inequality should hold:

\begin{equation}
\dfrac{1}{|C_i|-1} \sum_{y\in C_i} d(x,y) \leq \dfrac{1}{|C_j|} \sum_{y\in C_j} d(x,y), ,,, i\neq j.
\end{equation}
\end{defn}

The intuition behind the above notion is to ensure every sample point is associated with a cluster that has the highest average similarity.


\subsubsection{Unfairness Mitigation Algorithms}

\paragraph{Pre-processing Strategies} 

{\bf Fairlet Decomposition.} The concept of fairlets, which was introduced in~\cite{chierichetti2017fair}, aims to improve the fairness of various clustering algorithms based on the Balance metric. The strategy involves dividing the data points into small groups, or so-called fairlets, before performing clustering in such a way that the disparate impact doctrine is maintained. Each fairlet is then represented by a single point, and a vanilla clustering algorithm is applied to these representative points. Because the representative points are reasonably fair, the final clustering also tends to be fair. In~\cite{chierichetti2018fair}, near-linear algorithms are proposed for fairlet decomposition. Ahmadian et al.~\cite{ahmadian2020fair} employ the idea of fairlets for hierarchical clustering considering several objective functions such as revenue, value and cost. 


{\bf Data Augmentation.} Inspired by the approach in~\cite{rastegarpanah2019fighting}, Chhabra et al.~\cite{chhabra2022fair} propose data augmentation as an efficient method for fair clustering. The method involves augmenting the dataset using a small subset of data to achieve a fairer clustering output after applying the algorithm. The authors propose a general bi-level formulation to address two problem settings: 1) using convex group-level fairness notions and convex center-based clustering objectives, and 2) using general group-level fairness notions and general center-based clustering objectives.

\paragraph{In-processing Strategies} 

{\bf Regularizers and Constraints.} Built on the Balance metric, the authors in~\cite{kleindessner2019guarantees} incorporate fairness constraints in spectral clustering as well as providing empirical evidence that it is possible to achieve higher demographic proportionality at minimal additional cost in the clustering objective. Li et al.~\cite{li2020deep} introduce a fairness-adversarial term encouraging soft assignments that remain constant across various protected subgroups, resulting in a model that is not influenced by sensitive attributes. Zhang et al.~\cite{zhang2021deep} propose an approach for fairness in deep clustering. A regularization term based on the Balance notion is proposed and is combined with the clustering objective. Chai et al~\cite{chaifair} propose to use Sinkhorn divergence to reduce differences in predicted soft labels among various demographic groups and to develop representations that are conducive to clustering. The requirement of equalized confidence is modeled as a regularization term during training using Sinkhorn divergence, with several additional regularizers for ensuring accuracy.

{\bf Alternating Objective.} This approach focuses on entirely alternating between fairness and clustering objectives during unsupervised learning. Liu et al.~\cite{liu2023stochastic} formulate the cost of clustering and fairness as a bi-objective optimization problem to achieve balance. Their approach is based on mini-batch $k$-means clustering, where the algorithm performs clustering based on the $k$-means during mini-batch updates. However, the algorithm also includes a series of swap-based steps to enhance the balance in clusters. The routine involves exchanging data points between the least balanced and well-balanced clusters after the mini-batch update. The method described in~\cite{ziko2021variational} combines the Kuulback-Leibler fairness term with the objective of center-based and graph-based algorithms. The approach involves conducting a separate update for the assignment of datapoints based on the fairness objective and clustering objective. The heuristic algorithm proposed in~\cite{chen2019proportionally} focuses on achieving the proportionality fairness notion by alternating between fairness and clustering objective. The proposed algorithm achieves a $(1+\sqrt{2})$-proportional solution. 

\paragraph{Post-processing Strategies} The majority of the methods proposed for the post-processing stage involve using linear programming to reassign data points according to fairness metrics. This approach is referred to as the LP formulation. Additionally, Simoes et al.~\cite{simoes2022exploring} have recently explored an alternative approach based on {\em Data Perturbation} for fair clustering. The core concept of this approach is to perturb the assignment of data points to clusters in several iterations based on the "Rawls' difference principle"~\cite{altham1973rawls}.


{\bf LP formulation.} Considering disparate impact doctrine, the authors in~\cite{bera2019fair} show that for a given clustering with $l_p$-norm objective including center-based approaches such as $k$-means and $k$-medoids, it is possible to have fair algorithms with a slight sacrifice in fairness constraint. In more detail, given any $\rho$-approximation algorithm for a given clustering objective, a $(\rho+2)$-approximation solution exist for the best clustering, which satisfies fairness constraints. The objective is achieved by formulating and solving an LP optimization problem for fair assignment after clustering. Approaches in~\cite{ahmadian2019clustering} and~\cite{harb2020kfc} also use alternative LP formulations for fair assignment of datapoints to centers. Esmaeili et al.~\cite{esmaeili2020probabilistic} extend the approach to a scenario where data points are probabilistically assigned to groups instead of having a priori information on the group assignment. In~\cite{esmaeili2021fair}, first the center-based clustering algorithm is applied to maximize the clustering objective. Then, using an LP formulation, clustering is improved considering the fairness objective. This is done by searching for the cluster with maximum violation of the fairness objective considering the upper bound required for the clustering objective and rounding the possibly fractional solution to a feasible integer solution using a network flow algorithm.


\


\subsection{Fairness in Reinforcement Learning}
RL is concerned with learning how to make decisions in an environment by maximizing some cumulative reward (or equivalently, minimizing some regret)  through interacting with the environment and receiving feedback. The decisions made by RL agents can have a significant impact on individuals and society, making it essential to ensure that these decisions are unbiased and fair. Compared to other methods in which only the immediate impact of the decision-making algorithm is studied to mitigate unfairness, algorithms for fair RL aim to account for the long-term consequences of the agent's actions to ensure that they are unbiased \cite{wang2023survey}. 

In the context of fairness in RL, a significant focus is dedicated to addressing unfairness across various variants of the bandit problem. Bandit problems involve an agent repeatedly selecting from a set of arms, each associated with unknown reward distributions. The agent's objective is to determine an optimal \emph{policy} that maximizes cumulative reward while receiving limited feedback on unchosen arms \cite{lattimore2020bandit}. Bandit scenarios provide a tractable framework for exploring and developing fair decision-making algorithms in RL.


\subsubsection{Notions and Definitions}

In recent years, several perspectives have emerged for evaluating and improving the fairness of RL algorithms. Specifically, researchers have proposed and evaluated notions of fairness in RL from three distinct perspectives: \emph{Meritocratic Fairness}, \emph{Individual Fairness}, and \emph{Proportional Fairness}. These perspectives can be quantified in different ways, depending on the specific goals and objectives of the RL algorithm. 


{\bf{Meritocratic Fairness.}}
This perspective requires avoiding favoring less qualified individuals over more qualified ones. For example, in the context of bandits, this fairness notion indicates that it is unfair to preferentially select an arm with a lower expected reward over other available arms with higher expected rewards~\cite{joseph2016fairness}. This ensures that the rewards are allocated fairly based on the arms' abilities. The following definitions are some examples of evaluating fairness from this aspect in literature.

\begin{defn} \label{def: delta-fairness classic bandits}
($\delta$-Fairness in Classic Bandits~\cite{joseph2016fairness}).
An algorithm $\mathcal{A}$ is deemed $\delta$-fair if, with a probability of at least $1-\delta$ over history $h$, for all distributions $\mathcal{D}_1,...,\mathcal{D}_k$, every $t \in [T]$, and all $j,j' \in [k]$:
\begin{equation}
\pi^t_{j|h} > \pi^t_{j'|h} \text{ only if
} \mu_j > \mu_j'
\end{equation}
Here, $T$ represents a known horizon, $[k]={1,...,k}$ denotes the set of arms, and $\mathcal{D}_1,...,\mathcal{D}k$ are the unknown reward distributions of arms. $\mu_i$ is the unknown average reward of the $i$-th arm, and $\pi_{i|h}^t$ is the probability that $\mathcal{A}$ selects arm $i$ given history $h$.
\end{defn}
In essence, this definition suggests that selecting one arm over another is considered unfair if there is sufficient confidence to indicate that the chosen arm has a lower expected reward compared to the unselected one.

\begin{defn} \label{def: delta-fairness contextual bandits}
($\delta$-Fairness in Contextual Bandits~\cite{joseph2016fairness}).
An algorithm $\mathcal{A}$ is considered $\delta$-fair if, with a probability of at least $1-\delta$ over history $h$, for all sequences of contexts $x^1,...,x^t$, all payoff distributions $\mathcal{D}_1^t,...,\mathcal{D}_k^t$, every round $t \in [T]$, and all pairs of arms $j,j' \in [k]$:
\begin{equation}
\pi^t_{j|h} > \pi^t_{j'|h} \text{ only if
} f_j(x_j^t) > f_{j'}(x_{j'}^t),
\end{equation}
where $f_i:x_i \rightarrow [0,1]$ denotes an unknown mapping from the context to the reward for each arm.
\end{defn}

{\bf{Individual Fairness.}}
Individual fairness mandates that similar individuals should be treated similarly~\cite{dwork2012fairness}. In the context of bandits, it means that arms with similar qualities should be selected by the algorithm with similar probability~\cite{liu2017calibrated}. This ensures that no particular arm is consistently preferred over others that have similar expected rewards. The following definition provides a notion of individual fairness in the context of bandits.

\begin{defn} \label{def: smooth-fairness bandits}
"(Smooth Fairness~\cite{liu2017calibrated}). For a divergence function $D$, let $D(\pi_t(i) \mathbin\Vert \pi_t(j))$ denote the divergence between Bernoulli distributions with parameters $\pi_t(i)$ and $\pi_t(j)$, and let $D(r_i \mathbin\Vert r_j)$ denote the divergence between the reward distributions of the $i$-th and $j$-th arms. An algorithm $\mathcal{A}$ is $(\epsilon_1, \epsilon_2, \delta)$-fair with respect to the divergence function $D$ if $\epsilon_1, \epsilon_2 \geq 0$, and $0\leq \delta \leq 1$. With a probability of at least $1-\delta$ in every round $t$, for every pair of arms $i$ and $j$, the following inequality should hold:

\begin{equation}
D(\pi_t(i) \mathbin\Vert \pi_t(j)) \leq \epsilon_1 D(r_i\mathbin\Vert r_j) + \epsilon_2.
\end{equation}
\end{defn}
In other words, if two arms have comparable reward distributions, a fair decision rule should treat them similarly by assigning them similar selection probabilities.

{\bf{Proportional Fairness.}}
Proportional Fairness in RL aims to ensure that each user, or in the case of bandits, each arm, is allocated a minimum guaranteed share of resources or pulls over time \cite{li2019combinatorial, claure2020multi}. By doing so, it guarantees that each arm is played at least a certain proportion of the time, thus ensuring a minimum level of exploration for all arms and preventing any particular arm from being unfairly favored over others. We present the following fairness notion as a way to quantify Proportional Fairness:

\begin{defn} \label{def: asymptotic fairness}
(Asymptotic Fairness~\cite{li2019combinatorial}). 
Let $d(t) = (d_1(t), \ldots, d_N(t))$ be a vector indicating whether each of the $N$ arms is pulled at round $t$, where $d_i(t)=1$ if arm $i$ is played and $d_i(t)=0$ otherwise. Moreover, let $r_i \in (0, 1)$ denote the required minimum fraction of rounds in which arm $i$ is played. Algorithm $\mathcal{A}$ is called asymptotically fair if:
\begin{equation}
\liminf_{T \to \infty} \frac{1}{T} \sum_{t=0}^{T-1} \mathbb{E}[d_i(t)] \geq r_i, \forall i \in [N].
\end{equation}
\end{defn}
This means that as the number of rounds $T$ approaches infinity, the expected fraction of rounds in which each arm is played should be equal to or greater than a prespecified fraction that is considered fair.

\subsubsection{Unfairness Mitigation Algorithms}

Owing to the inherent characteristics of RL methods, most techniques in the ML category are in-processs. We have classified the mitigation algorithms into three groups: Reward modification, action constraint, and algorithmic adjustments.

{\bf Reward modification.}  Methods in this category involve modifying the notion of regret or rewards that the agent receives during training to encourage fairness. For instance, in \cite{patil2021achieving}, researchers propose an extension to the conventional notion of regret, called \emph{$r$-regret}, which incorporates fairness constraints to ensure that each arm is selected at least a pre-specified fraction of the time at each time step in stochastic multi-armed bandit problems. In a separate study on Interactive Recommender Systems, a novel RL-based framework named FairRec is introduced to combine accuracy and fairness in the rewards function for making recommendations \cite{liu2020balancing}. FairRec dynamically balances accuracy and fairness by incorporating user preferences and system fairness status into its state representations, which helps to improve fairness while preserving recommendation quality over time. 

{\bf Action constraint.} 
This category encompasses techniques that either constrain an agent's actions or adjust the action selection strategy during training to promote fairness. Joseph et al. \cite{joseph2016fairness} introduce \emph{FAIRBANDITS} for achieving $\delta$-fairness in stochastic bandits by modifying the UCB algorithm. When confidence intervals overlap, they recommend playing corresponding arms with equal probability. They also present a method for fairness in contextual bandit problems by converting the KWIK algorithm to a $\delta$-fair contextual bandit algorithm and vice versa. Jabbari et al. \cite{jabbari2017fairness} extend $\delta$-fairness to MDPs, ensuring no action is favored if it results in lower long-term discounted rewards. Their Fair-E\textsuperscript{3} algorithm achieves fairness based on an approximate definition. Liu et al. \cite{liu2017calibrated} propose the notion of \emph{smooth fairness}, a constraint based on the reward distributions as described earlier, and \emph{fairness regret}, to measure calibration deviations. They show how to address these constraints in Bernoulli and Dueling bandit settings.

{\bf Algorithmic Modifications.} This category involves modifying RL algorithms themselves to promote fairness. Some research in this area aims to guarantee a minimum number of times each arm is chosen in multi-armed bandit problems. For example, Chen et al. \cite{chen2020fair} incorporate constraints to ensure a minimum selection rate for each arm in contextual multi-armed bandit problems, suggesting an algorithm that minimizes regret for multiple contexts while maintaining fairness. Other studies target group-level fairness in RL-based decision-making. Huang et al. \cite{huang2022achieving} frame personalized recommendations as a modified contextual bandit problem, introducing a fair algorithm called \emph{Fair-LinUCB} to maintain parity in the expected mean reward of both the protected and unprotected groups. Wen et al. \cite{wen2021algorithms} propose fair sequential decision-making algorithms in MDPs that enforce fairness constraints based on average outcome quality for different subpopulations, aiming for demographic parity and equalized opportunity.

A key challenge in individual fairness is quantifying individual similarity. While many studies assume such a metric, Gillen et al. \cite{gillen2018online} propose learning a similarity metric during decision-making in contextual bandits setting, relying on an oracle that can identify fairness violations without explicitly providing a quantitative metric. 
Ge et al.~\cite{ge2022toward} introduce \emph{MoFIR}, a framework that balances fairness and utility in recommendation systems. The authors use Multi-Objective Reinforcement Learning to learn an optimal recommendation policy. MoFIR extends the Deep Deterministic Policy Gradient algorithm by incorporating a conditioned network that considers decision-maker preferences and outputs Q-value vectors. This approach aims to address fairness concerns while maximizing the effectiveness of recommendations.



\section{Privacy \& Fairness}\label{Sec: privacy vs fairness}

Privacy-focused methods such as DP aim to make individuals unidentifiable to outside observers, while fairness mechanisms aim to ensure balanced outputs across different groups. There are two different perspectives on the relationship between these two goals. One perspective sees them as compatible, while the other highlights a potential trade-off between them. In this section, we aim to explore this relationship in order to facilitate further research on more advanced algorithms that can achieve both goals simultaneously. The section is structured as follows:

\vspace{-6pt}
\begin{itemize}
    \item {\bf Architectures.} This subsection provides some common architectures that can be used to implement both privacy and fairness objectives together.
    \item {\bf Impact of Privacy on Fairness.} This section examines the empirical evidence on how achieving privacy can affect fairness.
    \item {\bf Impact of Fairness on Privacy.} This section focuses on the consequences and benefits of achieving fairness on privacy and presents existing evidence.
    \item {\bf Concurrent Implementation of Privacy and Fairness.} Here, we discuss algorithms that aim to achieve both privacy and fairness at the same time.
    \item {\bf Applications.} This subsection provides examples of the interrelation between privacy and fairness in various application domains.
\end{itemize}

\vspace{-6pt}


\subsection{Architectures}

\begin{figure*}[tb]
	\subfloat[]{%
	\includegraphics[scale = 0.35]{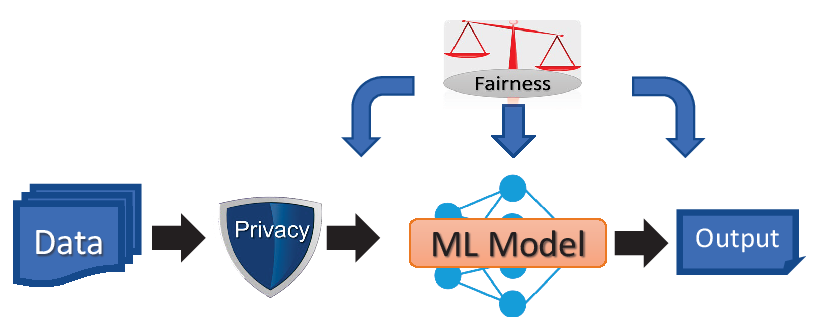}
	}
	\hfill
	\subfloat[]{%
	\includegraphics[scale = 0.35]{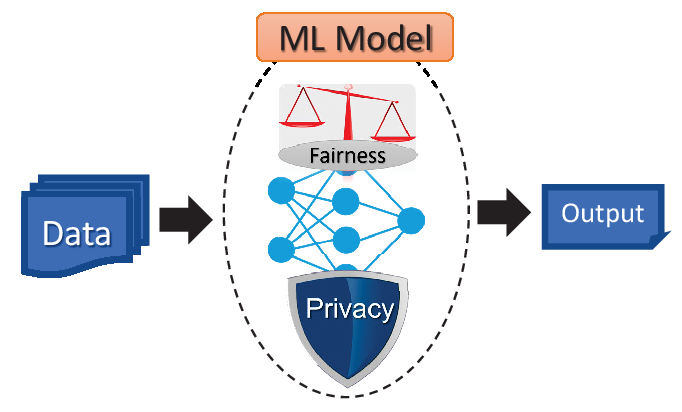}
	}
	\hfill
	\subfloat[]{%
	\includegraphics[scale = 0.35]{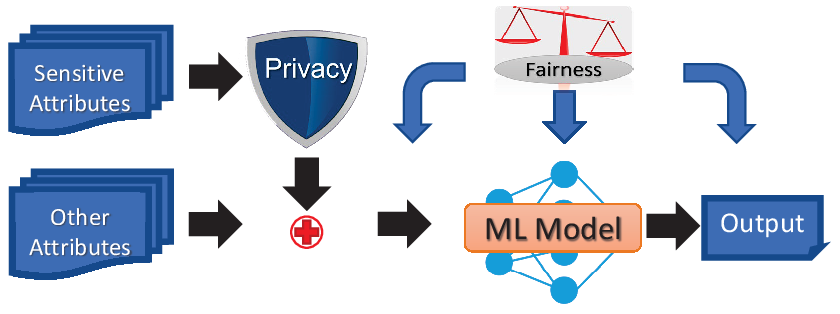}
	}
    \hfill
 	\subfloat[]{%
	\includegraphics[scale = 0.32]{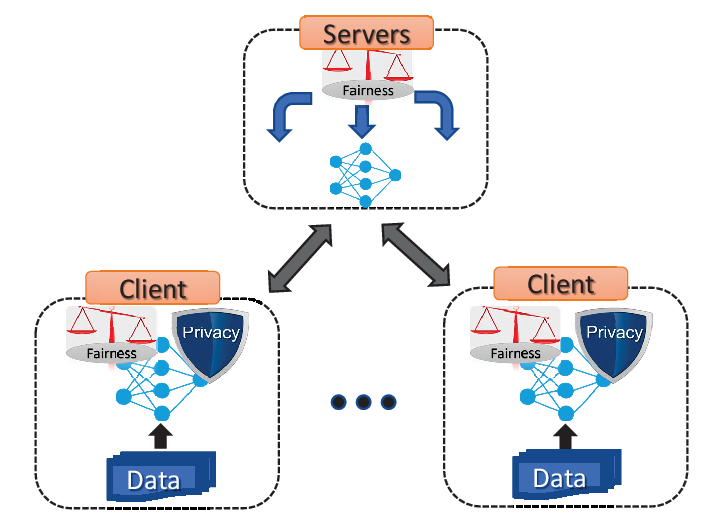}
	}
     \hspace{30pt}
	\subfloat[]{%
	\includegraphics[scale = 0.35]{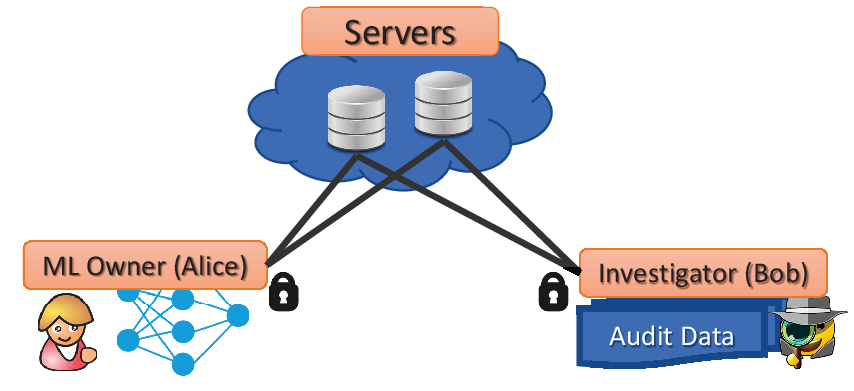}
	}
	\vspace{-10pt}
	\caption{Architectures Enabling Simultaneous Implementation of Privacy and Fairness in ML.}
	\label{Fig: Architecture}
	\vspace{-10pt}
\end{figure*}

In this section, we introduce five prominent architectures specifically developed to tackle the dual challenges of privacy and fairness in ML. The visual representations of these architectures are presented in Figure~\ref{Fig: Architecture}.

{\bf Architecture A.} In the first approach, a single entity seeks to ensure both privacy and fairness for their models. Initially, privacy-preserving algorithms are applied to the data, resulting in a privacy-protected dataset. This sanitized data is then used for fair training, with fairness pre-processing, in-processing, and post-processing techniques being applicable. For instance, the model-agnostic privacy-preserving k-means algorithm~\cite{su2016differentially} follows this approach by applying DP to data points before implementing the k-means algorithm. Techniques like fairlet decomposition~\cite{chierichetti2017fair} can be employed after sanitization to enhance fair learning.

{\bf Architecture B.} The second approach aims to achieve privacy and fairness simultaneously during model training. In this case, the ML model receives the data directly and attempts to perform fair and private learning, often using an augmented objective function, adversarial learning, or incentivizing methods in RL. An example of such an approach is seen in~\cite{chen2022fairness} and~\cite{liu2020fair}, where the authors focus on attaining objectives during training.

{\bf Architecture C.} The third approach involves separating sensitive and non-sensitive user attributes in the dataset. Sensitive user attributes undergo privacy-preserving methods for sanitization, after which both protected and unprotected attributes are passed through the ML pipeline. Fairness techniques can be applied at various stages, such as pre-, in-, or post-processing. This approach assumes the presence of a trusted entity with access to sensitive data, sharing them only in a private manner. An example of this method can be found in~\cite{lowy2022stochastic}.

{\bf Architecture D.} In the fourth scenario, an FL framework is highlighted, which leverages distributed learning. FL allows for the decentralized training of large-scale models without requiring direct access to clients' data, effectively maintaining their privacy. In a standard FL setting~\cite{mcmahan2017communication}, client nodes work together with a server to find a parameter vector that minimizes the weighted average of the loss across all clients. Techniques like Secure Aggregation~\cite{bonawitz2017practical} are used to guarantee that the server does not gain any information about the values of the individual updates sent by the clients, other than the aggregated value it aims to compute. Fairness is typically addressed through local debiasing methods and global unfairness mitigation strategies. A notable example of this structure is presented in~\cite{ezzeldin2021fairfed}.

{\bf Architecture E.} In the final common scenario, privacy-preserving fairness auditing is considered. We describe the framework in relation to Secure Multiparty Computation (MPC), a cryptographic technique that enables multiple parties to collaboratively compute a specific output from their confidential information in a distributed manner without actually exposing this private data. In this setup, a company (Alice) with a proprietary model must undergo an audit by an authority (Bob) with access to sensitive audit information. Alice wishes to keep her trained model parameters private, while Bob seeks to protect the sensitive audit data because it contains critical attributes necessary for fairness auditing but may also be subject to anti-discrimination and data protection regulations. A representative example of this privacy-fairness architecture is discussed in~\cite{pentyala2022privfair}.

\subsection{Impact of Privacy on Fairness}

The aim of this subsection is to comprehend the ways in which privacy-preserving methods affect the fair treatment of individuals and groups. The subsequent discussion reviews two perspectives on perspectives on the potential positive and negative impacts of privacy on fairness: one focused on ways in which they are aligned, and another focusing on ways in they contrast with each other.

\subsubsection{Aligned}

To begin our discussion, we examine publications that argue for the compatibility of privacy and fairness objectives. Pannekoek et al.~\cite{pannekoek2021investigating} build their experiments based on a simple neural network consisting of three fully connected layers. For fairness, they used the "reject option classification" approach as a postprocessing technique to adjust output labels. Meanwhile, for privacy protection, they implemented the DP version of the Adam optimizer~\cite{mcmahan2018general}. By incorporating both fairness and privacy constraints, their model displayed superior fairness compared to the model that only utilized fairness constraints, while maintaining high accuracy. Furthermore, the model's performance did not exhibit a declining trend as privacy protection measures increased. The authors in~\cite{khalili2021improving} demonstrate that the exponential mechanism developed to achieve DP can also help in attaining fairness once used as a post-processing approach at the output of classifiers in selection problems.

Sarhan et al.~\cite{sarhan2020fairness} examine the effect of DP on fairness in FL. The predominant scenarios in FL is explored including local DP and global DP. In the local DP scenario, clients use DP-SGD on their local nodes and transmit the model hyperparameters to the server for evaluation of fairness metrics such as equalized odds, equalized opportunity, and demographic parity. In contrast, in the global DP scenario, client hyperparameters are aggregated at the server and sanitized to achieve DP. The authors discover that DP reduces discrimination in both scenarios, but strict privacy budgets can result in a trade-off between privacy and fairness. As a result, the need for parameter tuning is emphasized.

The utilization of learned language models' hidden representations has been proven to be capable of extracting confidential information about users. In~\cite{lyu2020differentially}, the authors suggest a technique for safeguarding user privacy and demonstrate that incorporating privacy into the process can generally decrease bias. The proposed approach involves adding a noise layer to the extracted features from text, which achieves DP before sharing the representation for classification. Maheshwari et al. \cite{maheshwari2022fair} explore the integration of DP and adversarial learning, focusing on the Equalized Fairness metric in NLP. Their proposed framework perturbs the output of a text encoder to achieve DP, and employs a classifier branch in conjunction with an adversarial branch to actively foster fairness. The authors provide empirical evidence demonstrating that privacy and fairness are not only compatible in this context, but also mutually supportive.

\subsubsection{Contrasting}

On the contrary, some works argue that there exists a trade-off between privacy and fairness. Sanyal et al.~\cite{sanyal2022unfair} investigated the use of DP for classifiers, with a focus on accuracy discrepancy as a metric for fairness. Through their theoretical analysis, they established that it is not feasible to maintain high accuracy for minority groups and safeguard privacy simultaneously when data follows a long-tailed structure. Nevertheless, they demonstrated that relaxing accuracy requirements can lead to achieving a high degree of strict privacy and fairness. The authors in~\cite{bagdasaryan2019differential} focus on neural networks trained using Differentially Private Stochastic Gradient Descent (DP-SGD) and demonstrate that the accuracy of the private model drops more for the underrepresented classes. In other words, the DP-SGD amplifies the model’s bias towards popular elements in the distribution being learned. The results are empirically shown in gender classification, sentiment analysis of tweets,  species classification, and FL of language models.

Du et al.~\cite{du2019robust} study the role of DP on outlier detection. To detect outliers, an ML model is trained to learn the distribution from which the training data samples were drawn. Using this information, the model can detect samples that significantly differ from the learned distribution. The authors demonstrate that the random noise added during the model training for DP purposes hides the impact of an individual record on the learned model. Essentially, rare training examples are hidden by the added noise, resulting in a model that is less capable of detecting outliers. In other words, the accuracy of the model decreases for minority instances when DP is applied to the training process. In \cite{chen2022fairness}, the author validates the presence of a trade-off between privacy and fairness in semi-private settings, where a small fraction of sensitive data is clean, while the remaining data is safeguarded by privacy measures.

\subsection{Impact of Fairness on Privacy}



Despite the considerable impact that fairness may have on privacy, there's been relatively scant attention devoted to understanding how algorithmic fairness might affect the privacy of individuals and groups. Many strategies developed to tackle unfairness depend on sensitive user information, which can result in unintended or intentional overexploitation of such data, thereby violating user privacy. More discussion and dialog is needed within the ML community to comprehend the privacy hazards linked to algorithmic fairness. This need has been emphasized in numerous studies, such as~\cite{andrus2022demographic} and ~\cite{strobel2022data}. In the following, we delve into the few studies that have been undertaken on this subject matter.

\subsubsection{Aligned} 

In this subsection, we discuss methods where the pursuit of fairness in ML models has led to a positive influence on user privacy. Starting with one of the most significant results in this regard, the authors in~\cite{dwork2012fairness} demonstrate that the concept of individual fairness is indeed a generalization of the DP notion under specific distance metric definitions. This finding is crucial because it enables the use of algorithms developed for individual fairness to positively impact privacy. More specifically, inspired by \cite{fioretto2022differential}, consider the individual fairness notion \cite{dwork2012fairness} described in Definition \ref{def: Individual location fairness}. In \cite{dwork2012fairness}, the authors reveal that the mapping function $\mathcal{M}: X\rightarrow Y$ satisfies $\epsilon$-DP given that the similarity distance metrics for the input and output are defined as follows for any two users $u$ and $v$:

\begin{align}
&d_{\mathcal{X}}(u,v) = \epsilon |\boldsymbol{u} \Delta \boldsymbol{v}|,\
& d_{\mathcal{Y}}(\mathcal{M}(u),\mathcal{M}(v))= \underset{y\in \mathcal{Y}}{\text{sup}} \left( \dfrac{P(\mathcal{M}(\boldsymbol{u} = y))}{P(\mathcal{M}(\boldsymbol{v} = y))} \right).
\end{align}

Here, $\boldsymbol{u} \Delta \boldsymbol{v}$ denotes the set difference between two inputs $\boldsymbol{u}$ and $\boldsymbol{v}$ of $\mathcal{X}$.

In a different perspective, Aalmoes et al.~\cite{aalmoes2022leveraging} concentrate on attribute inference attacks as a measure of privacy risk, where the objective is to deduce sensitive characteristics such as race and gender from the training data of an ML model. In this scenario, the model is perceived as a black box, with the adversary having access to the model's output predictions for any input query. The authors explore how fairness constraints applied during model training influence the attribute inference attack. They find that fairness algorithms, which enforce equalized odds, serve as an effective safeguard against attribute inference attacks without affecting the model's utility. Consequently, the goals of algorithmic fairness and sensitive attribute privacy are found to be in harmony.

\subsubsection{Contrasting}

Chang et al.~\cite{chang2021privacy} define the risk of privacy as the success of membership inference attack on a trained model. In such attacks an adversary observes the model predictions attempting to distinguish between members and non-members of the training set. The proposed attack is used to compare the information leakage of models trained with and without fairness constraints on different groups in their training data. The authors provide empirical evidence showing that fairness-aware learning has a disproportionate impact on the privacy risks of subgroups, creating a trade-off. This phenomenon is explained by the fact that fair models have a greater tendency to memorize data from unprivileged subgroups, which makes them more vulnerable to membership inference attacks.

In another work, Zhang et al.~\cite{zhang2023interaction} examine the interplay between privacy and fairness within the node classification of GNNs. They provide empirical evidence of the negative influence of individual node fairness on edge privacy. In this study, the individual fairness notion is applied as treating nodes comparably in classification, ensuring they receive identical service quality irrespective of their backgrounds. Privacy, on the other hand, is assessed through link prediction attacks between nodes, which reveal the connections between two nodes in a specific pair. The authors provide empirical evidence highlighting the adverse effect of individual fairness on privacy in this setting.

\subsection{Concurrent Implementation of Privacy and Fairness}


This section will examine algorithms that have been proposed in existing literature with the aim of achieving both privacy and fairness objectives while minimizing the overall loss of utility. Unfortunately, there is a limited number of research works that have theoretically investigated the interaction between these two objectives, such as those presented in references~\cite{cummings2019compatibility} and~\cite{khalili2021improving}. Cummings et al.~\cite{cummings2019compatibility} prove that it is not possible to simultaneously achieve DP and perfect fairness in terms of equalized odds while maintaining higher accuracy than a constant classifier. On the contrary, the authors in~\cite{khalili2021improving} demonstrate that this assertion is not applicable in selection problems. In non-selection problems, the goal is to minimize the expected loss across the entire population while ensuring fairness constraints. For instance, in a hiring scenario, all applicants who meet the classifier's criteria should be accepted. However, in selection problems, only a limited number of candidates can be chosen. The authors in~\cite{khalili2021improving} suggest that the exponential mechanism developed for DP can be an effective tool for improving fairness under specific circumstances.

Liu et al.~\cite{liu2020fair} develop an algorithm called FairDP, which aims to address disparate impact introduced by DP on underrepresented groups in the private training of classification models. The authors frame the learning process as a bilevel programming problem, which incorporates fairness and DP. FairDP utilizes an adaptive clipping threshold to regulate the impact of instances in each class, enabling the model accuracy to be adjusted for classes based on its privacy cost and fairness considerations. Chen et al.~\cite{chen2022fairness} investigated fair classification in semi-private settings where sensitive attributes, such as gender, are secured by specific privacy mechanisms, and only a few clean attributes are available. Their proposed framework aims to utilize the limited clean attributes to correct the noisy sensitive attributes while ensuring privacy.

Xu et al.~\cite{xu2019achieving} propose a method to combine privacy and fairness in logistic regression while maintaining high model accuracy. Their approach involves adding a decision boundary fairness constraint to the objective function and applying the functional mechanism for DP. The decision boundary fairness constraint is defined as the correlation between users' protected attributes and the distance from their unprotected attribute vectors to the decision boundary. The functional mechanism adds randomness to the polynomial coefficients of the constrained objective function by introducing Laplace noise. The experiments demonstrate a trade-off between the amount of privacy budget in DP and discrimination based on statistical parity. The authors in~\cite{hajian2015discrimination} argue for the benefits of addressing discrimination risk and privacy concerns in data mining, with a focus on the privacy metric $k$-anonymity. The authors consider a scenario where a set of patterns needs to be published in a way that preserves privacy and avoids discrimination, while minimizing pattern distortion. To achieve this goal, the proposed approach first identifies patterns that are more susceptible to vulnerability during the sanitization process, and then applies special measures to sanitize them based on the $k$-anonymity principle.

Jin et al.~\cite{jin2022privacy} examine an inference as service (IAS) scenario to make decisions in the cloud. In this setting, data is transmitted from devices to the cloud, and the cloud provider's server trains the ML model. To enhance user privacy and decrease bias, the author suggests a random mapping of data generated based on a non-convex optimization problem and an iterative algorithm to solve it. Inference accuracy, the mutual information between the transformed variable and the label, and the degree of leaked information are used to evaluate the utility, privacy, and impartiality of the mapping.

There are multiple works that focus on achieving privacy and fairness objectives simultaneously for ML models in the FL setting. Padala et al.~\cite{ padala2021federated} propose a framework for incorporating both privacy and fairness into FL. The authors use statistical parity as fairness metric and local DP as a means of ensuring privacy. The proposed framework is based on separating the training dataset and following a two-step process. In the first phase, each client trains a model on their own private dataset for unbiased prediction. In the second phase, clients train a DP model to mimic the unbiased prediction achieved in the first phase. Once training is complete, client data is transmitted to the server. Zhang et al.~\cite{zhang2020fairfl} propose a framework called FairFL to address the challenges of restricted information and constrained coordination in FL. The authors use statistical parity as a fairness metric and ensure privacy by not sharing raw training and ground truth labels, as well as sensitive demographic information. The FairFL framework consists of two components: a Team Markov Game for Client Selection (TMGCS) and a Secure Aggregation Protocol (SAP). The TMGCS is a Multi-Agent Reinforcement Learning approach that allows clients to collaboratively decide whether to participate in the local update process, while the SAP addresses the issue of each client's myopic view by allowing them to gather information about all clients' statuses without violating their privacy constraints.

\subsection{Applications}

In this section, we will examine domain-specific approaches that researchers have taken to incorporate privacy and fairness as key components of trustworthy ML in their respective applications.

{\bf Healthcare.} Patient electronic health records (EHRs) are regarded as some of the most sensitive information available, subject to stringent legal protection. However, such data is highly valuable to researchers and decision-makers, providing guidance for policy creation and medical advancements, as demonstrated during the recent unfortunate Covid-19 pandemic. This data often includes highly personal and sensitive patient information, such as gender, race, and age, which must be considered to ensure equitable treatment of individuals and groups. A prevalent solution for balancing privacy with the incorporation of fairness is utilizing synthetic datasets that mimic patient records to enhance privacy and train equitable models for future predictions~\cite{seastedt2022global}. Determining the degree of resemblance (privacy), fairness notions in healthcare, and their interplay remains an ongoing challenge and an active research area. For instance, in~\cite{bhanot2021problem}, multiple fairness notions have been suggested for covariate-level insights in synthetically generated healthcare data, underscoring the need for additional research to comprehend the interaction between privacy and fairness.

{\bf Natural Language Processing.} Modern natural language processing (NLP) models heavily rely on the encoded representation of text, which often captures sensitive attributes about individuals (e.g., race or gender). This raises privacy concerns and can cause downstream models to be unfair to certain groups. Sensitive information can be implicitly or explicitly present in the input text. Maheshwari et al.~\cite{maheshwari2022fair} propose a framework to address this problem. The framework aims to perturb the output of a text encoder to achieve DP and then uses a classifier branch along with an adversarial branch to actively promote fairness. The authors show that not only are these two objectives not in conflict with each other, but they also help each other in improving both objectives.  The proposed approach in~\cite{lyu2020differentially} adds a noise layer to the feature extraction process, achieving DP before using the representation for classification tasks, also showing the alignment of two objectives.

{\bf Computer Vision.} Comprehending privacy in Computer Vision involves various facets; however, present approaches for addressing privacy and fairness interplay mainly focus on defining privacy as instances where the model might memorize the training data in some way or use features as proxies for the original data to deduce private attributes about individuals, a concept known as attribute privacy. In this case, the model inadvertently or intentionally transfers information about the private attribute into the features, bearing resemblance to fairness-related concerns. The authors in~\cite{paul2023evaluating} examine the privacy-fairness trade-off, concentrating on the method where a feature extractor transforms images into features using weights. They employ an adversarial technique to integrate privacy and fairness by adding penalty terms for both objectives to the feature extractor's loss function during training. Findings reveal that privacy and fairness are conflicting in this context. In~\cite{tian2022fairness}, facial attribute classification is explored. The authors leverage GANs to maintain privacy through the generation of synthetic images and use contrastive learning-based loss designs to concurrently enforce fairness protections.

{\bf Spatial Data Processing.} User location data contains essential information about individuals, such as socio-economic aspects and indicators for sensitive attributes like race. With the US 2020 census data being published using DP across neighborhoods, comprehending the relationship between privacy and fairness has grown increasingly crucial. In~\cite{pujol2020fair}, the authors explore the impact of DP on supervised decision-making in three applications: (I) assigning voting rights benefits to minority languages, where privacy noise leads to significant disparities in accurately identifying deserving beneficiaries; (II) parliamentary apportionment, where certain privacy budget settings result in a more equitable distribution of seats to Indian states using noisy data compared to deterministic methods; and (III) federal funds allocation, where under strict privacy settings, some districts receive an uneven share of funds. The study underscores DP applied to location data can lead to disparities in the outcome, emphasizing the need for more in-depth research on privacy's influence on fairness. Furthermore, fairness in spatial data is a relatively new concept, with studies like~\cite{shaham2022models} and~\cite{shaham2023fair} addressing issues such as discrepancies in fairness metrics across neighborhoods, abrupt changes in classifier outputs due to neighborhood shifts, and the use of spatial continuity to enhance fairness. The impact of privacy on these aspects has not yet been considered and calls for further investigation.

\section{Vision and Challenges}\label{Sec: Vision and Challenges}






\subsection{Privacy and Fairness in the Current Era of Large Language Models}

As we step into the era of large language models, the interplay between privacy and fairness becomes an ever more critical component. As models tend towards being more conversational, they will exhibit higher risks of running into privacy and fairness violations. There are several avenues of active research that might dictate the future of this area.


{\bf The Use of APIs.} A significant number LLMs have restricted access \citep{brown2020language, openai2023gpt4, Soltan2022} and some of these models can be accessed via APIs. These APIs can host models along with an arsenal of ex-ante and post hoc qualitative checks that enable API owners to control privacy as well as fairness of the model output. These qualitative checks are implemented via a variety of techniques ranging from simple filters to secondary models that are trained to detect and process model output in accordance with some chosen policies. As LLMs and related APIs become more ubiquitous, the qualitative checks implemented by these APIs are likely going to be key in terms of mitigating bias and privacy issues and improving the overall responsibility of model output. 


{\bf Logic-aware Models.} When it comes to building models that are fair, the de-biasing process can have an effect on privacy preservation \citep{Agarwal2020TradeOffsBF}. One way to approach this problem is explore bias mitigation methods that skip this step. Logic-aware language models \citep{luo2023logic} can be trained to reduce harmful stereotypes. Instead of typical sentence encoding they use a textual entailment which learns if parts of the second sentence text {\it entails}, {\it contradicts} or is neutral with respect to parts of the first one. Models trained in this way were significantly less biased than other baselines, without any extra data or additional training paradigms used. Logic-aware training methods might be paired with privacy preservation techniques in order to build models that are both private and fair. In order to address privacy-preservation, smaller (i.e. 500X smaller than the state-of-the-art models; \citep{luo2023logic}) logical language models that are qualitatively measured as fair, can be deployed locally with no human-annotated training samples for downstream tasks.


{\bf Privacy and Fairness in the Context of Learning from Human Feedback.} Fine-tuning with human feedback \citep{ouyang2022training}, has provided a promising way to make large language models align more with human intent. This technique can also be utilized to train models that are privacy-preserving and unbiased. Here, modes are expected to learn to return content that is preferred by humans, based on a training loop where feedback is provided via a reward model trained to rank model output based on what humans might prefer. The collection of human preferences that are used to train the reward model can be made in a fair and private way so that the reward model will learn these traits. This will in turn enable the foundational model to learn to generalize its behavior based on the feedback provided by the fair and privacy preserving reward model. It has already shown some promise in reducing harmful content \citep{ouyang2022training}, but more research in this area is needed.

\subsection{Fairness Through Privacy}

The majority of previous approaches aimed at mitigating bias require access to sensitive attributes. However, obtaining a significant amount of data with sensitive attributes is often impractical due to people's growing privacy concerns and legal compliance. Consequently, a crucial area of research inquiry that merits attention is how to ensure fair predictions while preserving privacy. This is a persistent challenge faced by technology companies that seek to balance the goal of ensuring fair ML processing of user data, including sensitive attributes such as Race and Gender, while simultaneously protecting user privacy and restricting the use of sensitive user data.


\subsection{Fair Privacy Protection}


The authors in~\cite{ekstrand2018privacy} pose a crucial question that sparked this subsection: Does a system provide equivalent privacy protections to different groups of individuals? The main idea behind fair privacy protection is to ensure that privacy mechanisms offer equal levels of privacy to all users, meaning that users are being treated fairly in terms of the amount of privacy protection they receive. Although there is a lower limit on the level of privacy achieved, such as in DP, some groups of the population may receive more attention than others in a broader context.

The significance of fair privacy protection is explained in the following example predicated on an observation made in DP publication of the US $2020$ census. In an observation made by the US Census Bureau Researchers, the Laplace mechanism in DP appeared to be disadvantaging low-populated areas like villages compared to highly populated cities such as metropolitan areas~\cite{o2019differential}. To demonstrate this, let us consider two cities, $A$ and $B$, with populations $a$ and $b$, respectively, where $a<<b$. The populations are sanitized using Laplace noise, with two noise values drawn from a Laplace distribution ($Lap(1/\epsilon )$) added to each population, and the private values are published. At first glance, both cities appear to be sanitized using the same Laplace distribution, and both achieve $\epsilon$-DP. However, upon closer inspection, the amount of noise added per individual in each city is examined. With knowledge that the variance of Laplace noise is $2/\epsilon^2$, the amount of noise variance per individual is derived as $2/(a\epsilon^ 2)$ and $2/(b\epsilon^2 )$. If $a<<b$, then it can be seen that $2/(b\epsilon^2 )$ is much less than $2/(a\epsilon^2 )$. In other words, the amount of noise per individual in the low-populated city is much higher than in the highly populated city, which raises questions about the fairness of the privacy guarantees imposed.



\subsection{Incorporating Privacy and Fairness based on Cryptographic Approaches}

No existing approach addresses {\em both} privacy and fairness in the cryptographic setting. Such an approach presents great promise, because it may be able to provide privacy and fairness under more relaxed system architecture assumptions. For instance, in the differential privacy case, one assumes the presence of a trusted curator, or that an extensive distributed infrastructure for federated learning exists. With cryptography, no trusted party is required to perform the computation.

The main challenge becomes how can one express fairness constraints so that they become implementable using the rather restrictive set of operations provided by various searchable encryption approaches. Can one achieve fairness directly under the encrypted ciphertext using primitives like PHE? Or are there more expensive primitives required, like FHE? And even with FHE, only polynomial evaluation is supported in the best case, whereas other operations (e.g., logarithm, sigmoid) must be simulated using polynomial approximations. Achieving fairness by using such primitives is an important and challenging research problem.


\section{Conclusion}\label{sec: conclusion}
In conclusion, this comprehensive survey offers a thorough investigation of the fundamental concepts in privacy and fairness by examining nearly $200$ works in the field. Our aim is to guide researchers in both academia and industry towards the simultaneous realization of privacy and fairness for individuals and groups in society across all four primary facets of ML, including supervised, unsupervised, semi-supervised, and reinforcement learning. By establishing a solid understanding of privacy and fairness within various ML techniques, we present an exhaustive analysis of how objectives impact one another and identify open questions for the first time. This work emphasizes the focus areas necessary to address the dual objectives in ML, ultimately promoting more responsible and trustworthy decision-making.\\


\setcounter{page}{35}


\appendix

\section{Mind Map of Survey}\label{section: mind map}

\begin{figure*}[htbp]
\centerline{\includegraphics[scale=0.24]{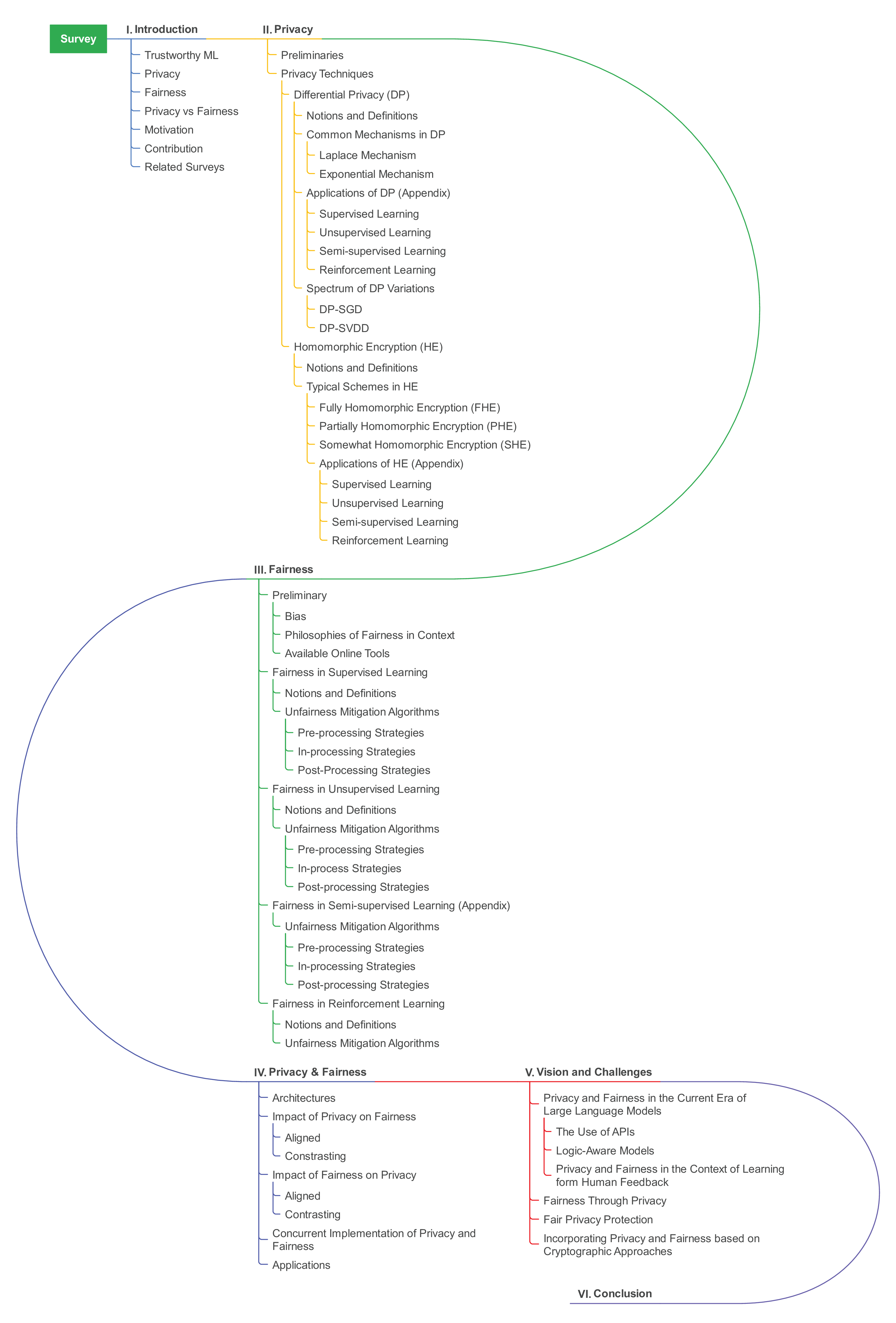}}
\label{fig: mind map}
\end{figure*}


\section{Applications of DP}\label{section: appendix applications of DP}

This section explores the practical use cases of DP in supervised, unsupervised, semi-supervised, and RL.

\subsection{Supervised Learning}
 Various studies have demonstrated the crucial role of DP in enhancing the privacy of ML tasks such as regression and decision tree classification. For instance, Sheffet et al. \cite{sheffet2019old} proposed DP algorithms for approximating the 2nd-moment matrix, while Milionis et al. \cite{milionis2022differentially} presented efficient DP algorithms for classical regression settings. Furthermore, Kim et al. \cite{kim2019secure} proposed a hybrid approach using DP and homomorphic encryption for privacy-preserving distributed logistic regression. In the context of decision tree classification, Liu et al. \cite{liu2018differentially} and Fletcher et al. \cite{fletcher2015differentially} \cite{fletcher2017differentially} introduced decision tree and forest algorithms that ensure privacy while achieving high accuracy. Their proposed algorithms showcase the importance of DP in the development of ML systems that maintain high levels of privacy without compromising performance. 
 \newline
Random Forest classification algorithms have also been shown to benefit significantly from DP techniques. Specifically, Patil et al. \cite{patil2014differential} and Hou et al. \cite{hou2019dprf} proposed algorithms that integrate DP with the Gini index and demonstrate the effectiveness of DP in preserving classification accuracy and privacy in Random Forest classification. Moreover, the significance of DP in SVM for safeguarding privacy in ML is exemplified by Senekane et al. \cite{senekane2019differentially} and Park et al. \cite{park2023efficient}. Senekane et al. proposed privacy-preserving image classification using SVM and DP with $\epsilon$-DP, while Park et al. proposed an algorithm for multi-class classification utilizing SVDD with DP-SVDD for training and EPs for classification, respectively. Their approaches showcase the efficacy of DP in enhancing the privacy of ML systems while maintaining high levels of accuracy.

\subsection{Unsupervised Learning}
Numerous investigations have explored the implementation of DP in unsupervised learning. For example, Tantipongpipat et al. \cite{tantipongpipat2019differentially} propound a conceptual framework that amalgamates autoencoder and GANs to create top-tier synthetic data in unsupervised settings. They also introduce novel metrics to appraise the excellence of synthetic data and showcase the efficacy of their approach on medical and census datasets. Similarly, Torfi et al. \cite{torfi2022differentially} confront the challenges of synthetic data generation in medical domains, such as safeguarding privacy, handling discrete data, and incorporating temporal and correlated features. They put forth a privacy-preserving framework that utilizes RDP-CGAN and surpasses current methodologies in terms of privacy guarantee and quality of synthetic data. Their approach is designed to provide both privacy and quality in synthetic data generation.
\newline
Other studies have focused on employing DP to particular unsupervised learning tasks. Specifically, Wang et al. \cite{wang2015differentially} scrutinize differentially private subspace clustering algorithms and present a pragmatic Gibbs sampling subspace clustering algorithm using the exponential mechanism. This algorithm is designed to preserve the privacy of individuals' data. In another study, Bun et al. \cite{bun2021differentially} introduce an algorithm for differentially private correlation clustering that accomplishes subquadratic additive error compared to the optimal cost. Their approach improves upon previous methods, which have typically incurred quadratic costs. Similarly, Blocki et al. \cite{blocki2021differentially} devise private sublinear-time clustering algorithms for k-median and k-means clustering in metric spaces, and initiate a sampling algorithm with group privacy analysis. Their approach focuses on reducing computation time while ensuring privacy. In yet another application, the authors in \cite{niinimaki2019representation} address privacy preservation in human genomic datasets and propose differentially private ML using representation learning, evincing enhanced accuracy in drug sensitivity prediction.

\subsection{Semi-supervised Learning}
Both \cite{pham2018differentially} and \cite{long2017differentially} offer novel frameworks for differentially private semi-supervised classification, utilizing both labeled and unlabeled data to train a classifier. Whereas the former prioritizes known class priors and provides proofs on privacy and utility, the latter introduces two distinctively differentially private methods, output perturbation and objective perturbation, and assesses their performance against regular differentially private empirical risk minimization (ERM). In addition, \cite{long2017differentially} conducts a thorough analysis of the global sensitivity of the objective function in SSL, demonstrating that their proposed methods attain superior accuracy while ensuring DP.
\newline
In contrast, \cite{jagannathan2013semi} tackles the issue of enhancing the accuracy of a differentially private classifier using non-private data when only a small amount of private data is accessible. Their approach fabricates a differentially private classifier from private data and subsequently employs non-private data to boost accuracy, extending the random decision tree idea to leverage the availability of unlabeled data for denser partitioning of the instance space and label propagation. This method also boosts classifier accuracy without compromising privacy, as demonstrated on small and moderate-sized datasets. This diverges from the differentially private boosting algorithm that amplifies the accuracy of a class of real-valued queries. All in all, these studies underscore the significance of DP in ML and propose compelling solutions to address privacy concerns while enhancing classification accuracy.

\subsection{Reinforcement Learning}
The studies \cite{ma2020differentially} and \cite{zhou2022differentially} make contributions to the field of privacy-preserving RL. Whereas \cite{ma2020differentially} proffers solutions to attain DP in RL contexts through the utilization of exponential and Laplace mechanisms, \cite{zhou2022differentially} posits a novel approach for devising privacy-preserving RL algorithms with rigorous statistical guarantees. The latter expounds both value-based and policy-based optimistic private RL algorithms under linear mixture MDPs, which revel in sublinear regret in the total number of steps while ensuring joint DP. These results open up new avenues for safeguarding data privacy while ensuring the scalability and efficiency of RL algorithms in large-scale MDPs.
\newline
Similarly, the studies \cite{cheng2022multi} and \cite{li2019differentially} advance approaches for integrating DP in meta-learning and multi-agent RL, correspondingly. \cite{cheng2022multi} ushers in a new technique for translocating knowledge with DP, christened Differential knowledge Transfer with relevance Weight, which boosts the model's resilience to negative transfer and augments the knowledge set. Meanwhile, \cite{li2019differentially} proffers a new framework for incorporating DP into meta-learning, with encouraging results in both theory and practice. Their proposed privacy setting accords better performance than previously studied notions of privacy. These studies make significant contributions to broadening the horizons of privacy-preserving ML algorithms in diverse learning scenarios.


\section{Applications of HE}\label{section: appendix applications of HE}
This section explores the practical use cases of HE in supervised, unsupervised, semi-supervised, and reinforcement learning.

\subsection{Supervised Learning}
HE has been applied to several supervised learning models to ensure privacy and protect sensitive data. One study proposed a secure method for linear regression on split data using Paillier's PHE scheme, which allows for the homomorphic addition of encrypted values \cite{hall2011secure}. Another study introduced novel techniques using FHE to train logistic regression models on encrypted data, with potential applications to other models such as neural networks \cite{chen2018logistic}. In addition, a homomorphically encrypted logistic regression outsourcing model was presented, allowing ML models to be learned without accessing raw data \cite{kim2018secure}. However, there are still limitations to the application of these models, including overheads in computation and storage due to HE, and the need for polynomial approximation to reduce computation cost.
\newline
HE has also been applied to decision tree and random forest models. One study proposed SortingHat, an efficient non-interactive design for private decision tree evaluation using FHE techniques \cite{cong2022sortinghat}. SortingHat addresses cryptographic problems related to FHE and presents a version without transciphering that significantly improves computation cost. Another study proposed a privacy-preserving protocol for multiple model owners to delegate the evaluation of random forests to an untrusted party, incorporating a SHE scheme and optimization techniques \cite{aloufi2019blindfolded}. Both studies present new secure protocols and optimization techniques to enable practical use of heavy-weight HE schemes. Finally, a study presented a secure multi-label tumor classification method using HE, based on a neural network model with the softmax activation function \cite{hong2022secure}. The model achieved high accuracy and successfully computed the tumor classification inference steps on encrypted test data.

\subsection{Unsupervised Learning}
HE has gained attention as a promising tool for privacy-preserving ML, especially in unsupervised learning tasks such as clustering and principal component analysis. Several studies have proposed different techniques to utilize HE for these tasks. For instance, Catak et al. \cite{catak2020practical} proposed a privacy-preserving clustering system that utilized Paillier Cryptography and HE to preserve privacy and minimize computational time. Alabdulatif et al. \cite{alabdulatif2017privacy} introduced a distributed data clustering approach using fully homomorphic encryption, resulting in significant improvements in computational performance efficiency for clustering tasks. Wu et al. \cite{wu2020secure} proposed a secure and efficient outsourced k-means clustering scheme using YASHE homomorphic encryption, achieving privacy preservation in database security, clustering results, and data access pattern hiding.
\newline
In addition, HE has also been applied to principal component analysis. Panda et al. \cite{panda2021principal} proposed a non-interactive technique to perform principal component analysis using the CKKS HE scheme, achieving good performance on higher-dimensional datasets with a sub-ciphertext packing technique that reduces computations. Moreover, HE has also been used for privacy-preserving recurrent neural networks. Bakshi et al. \cite{bakshi2020cryptornn} proposed privacy-preserving recurrent neural networks using HE, evaluating five different methods to deal with increasing noise and linearity of activation functions on various datasets and network architectures, and proposing three original methods to retain non-linear function benefits.

\subsection{Semi-supervised Learning}
HE has shown great potential in supporting SSL tasks by enabling secure computation on encrypted data. The study by Arai et al. \cite{arai2011privacy} proposes a novel privacy-preserving label prediction solution with HE, while the study by Erkin et al. \cite{erkin2012generating} uses secure multiplication and decryption protocols, as well as data packing, to provide a privacy-preserving recommender system. On the other hand, the study by Pejic et al. \cite{pejic2022effect} compares the performance loss of different HE techniques and Multi-Party Computations (MPC) in Federated Learning (FL) to train a Generative Adversarial Network (GAN) on sensitive data.
\newline
Despite their differences in applications and techniques, all three studies highlight the potential of HE in protecting sensitive data while still allowing for useful computations to be performed. Arai et al. \cite{arai2011privacy} and Erkin et al. \cite{erkin2012generating} demonstrate the feasibility and efficiency of privacy-preserving solutions with HE in their respective fields, while Pejic et al. \cite{pejic2022effect} show the trade-off between the complexity of encryption methods and the time taken for computations. These studies suggest that HE can enable SSL tasks with privacy concerns in various contexts.

\subsection{Reinforcement Learning}
HE has become an increasingly popular tool for addressing privacy and security concerns in RL in cloud computing and IoT environments. Park et al. \cite{park2020privacy} proposed the Secure Q-Learning algorithm using FHE, which processes data in a single cloud server and restricts error growth without the bootstrapping algorithm. In addition, Suh et al. \cite{suh2021sarsa} developed the encrypted SARSA(0) algorithm, which offers privacy guarantees and induces minimal precision loss in control synthesis over FHE. Both studies demonstrate the potential of HE in addressing security and privacy challenges in RL tasks. 
\newline
Moreover, Miao et al. \cite{miao2021federated} proposed a FL based Secure data Sharing mechanism (FL2S) for IoT with privacy preservation, which uses an asynchronous multiple FL scheme with sub-task grading and deep RL, as well as HE for privacy protection. Meanwhile, Sun et al. \cite{sun2021privacy} proposed secure computation protocols using FHE for the A3C RL algorithm in health data, and design the first secure A3C RL algorithm for treatment decision-making. These studies demonstrate the efficiency of HE-based protocols and algorithms in secure RL and offer promising solutions to privacy and security concerns in various domains.

\section{Bias}\label{section: appendix bias}

\begin{figure}[t]
\includegraphics[scale=.5]{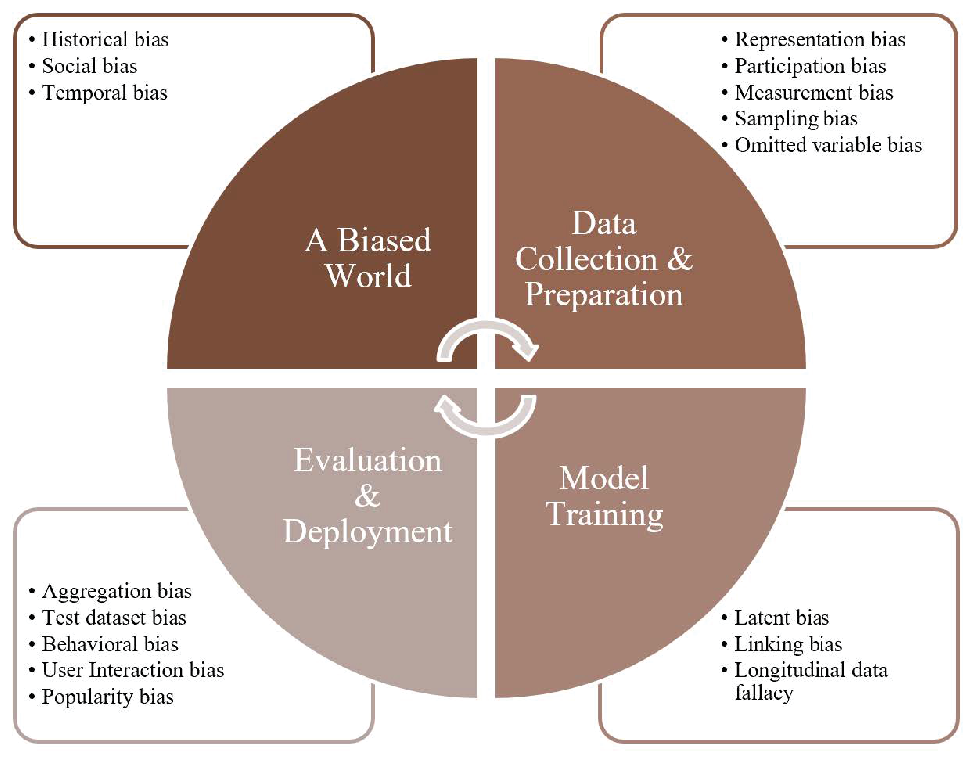}
\centering
\caption{Categorization of bias types in ML.}
\vspace{-15pt}
\label{Fig: categorization of bias}
\end{figure}

This section presents our classification of the different forms of bias illustrated in Figure~\ref{Fig: categorization of bias}. The types of bias have been divided into four major categories, namely, \emph{A Biased World}, \emph{Data Collection and Preparation}, \emph{Model Training}, and finally, \emph{Evaluation and Deployment}.

{\bf A Biased World.} Even if all fairness-related considerations are made for ML models, a significant source of bias is introduced in the data cycle due to perceptions and beliefs in society. \emph{Historical bias}~\cite{suresh2019framework} is the primary source in this category that highlights the adverse effects of reflecting historically unfair facts in ML models. For example, in 2018, image search results for female CEOs showed a bias towards male CEOs, reflecting the underrepresentation of women as CEOs in Fortune 500 companies (5 only)~\cite{suresh2019framework}. The question of whether search algorithms should reflect this reality remains uncertain~\cite{mehrabi2021survey}.

 {\bf Data Collection \& Preparation.} The role of data and how they are transformed is pivotal in ensuring fairness practices are in place. Well-known sources of bias in this category include {\em Representation bias} occurring when certain parts of input space are not reflected in the collected data, {\em Measurement bias} due to intentional or unintentional use of certain features as proxies for others, particularly for sensitive attribute, and {\em Sampling bias} in which non-random sampling of the dataset is used for the purpose of training~\cite{zadrozny2004learning}.

{\bf Model Training.} The most subtle but equally detrimental sources of unfairness are introduced during training. {\em Latent bias} is one such factor for which the model will learn existing biases in society. For example, existing stereotypes in society are learned and reflected by the model. Another such bias is called {\em Linking bias} caused by learning fundamentally different patterns from real-world counterparts raised mostly in models applied on social networks~\cite{bakshy2012role}.

 {\bf Evaluation \& Deployment.} The most intuitive types of bias incur in the evaluation and somewhat less obvious ones in the deployment of ML models. For example, an unrepresentative test dataset is formulated under {\em Test Dataset bias} incurred during evaluation, and {\em Behavioral bias} formulates systematic distortions in user behavior across platforms or contexts, or across users represented in datasets~\cite{olteanu2019social}. Miller et al.~\cite{miller2016blissfully} demonstrated behavior bias by considering types of emojis existing on different platforms and how they can lead to different user reactions.

\section{Sensitive Attributes}\label{section: appendix sensitive attributes}

Protected attributes are characteristics like gender and race that are determined by either legal obligations or specific values of an organization. Laws like the US Fair Housing Act, the Federal Equal Employment Opportunity, and the equal credit opportunity act specify requirements for algorithmic fairness in areas such as housing, employment, and credit. The following table briefly summarizes protected attributes based on the particular organization.

\begin{table}[]
    \caption{Sensitive attributes in different organizations.}
\renewcommand{\arraystretch}{1.35}
\begin{tabular}{|c|p{2.9cm}|p{3.5cm}|p{3.5cm}|}
\hline
\textbf{Attribute} & \textbf{US Fair Housing Act} & \makecell[c]{\textbf{Employment}\\\textbf{ Opportunity Act}} & \makecell[c]{\textbf{Equal Credit}\\\textbf{ Opportunity Act}} \\ \hline 

Race or Color  & \hfil \checkmark & \hfil \checkmark  & \hfil \checkmark \\\hline

Sex & \hfil \checkmark & \hfil \checkmark  & \hfil \checkmark \\ \hline

Religion  & \hfil \checkmark & \hfil \checkmark & \hfil \checkmark  \\ \hline

National   Origin & \hfil \checkmark  & \hfil \checkmark & \hfil \checkmark \\ \hline

Marital   Status &  \makecell[c]{$\times$} & \makecell[c]{$\times$} & \hfil \checkmark \\ \hline

Familial   Status & \hfil \checkmark & \makecell[c]{$\times$} &   \makecell[c]{$\times$} \\ \hline

Disability & \hfil \checkmark & \hfil \checkmark &  \makecell[c]{$\times$} \\ \hline

Age & \makecell[c]{$\times$} & \hfil \checkmark & \hfil \checkmark \\ \hline

Genetic   Info & \makecell[c]{$\times$} & \hfil \checkmark & \makecell[c]{$\times$} \\ \hline

Pregnancy & \makecell[c]{$\times$} & \hfil \checkmark & \hfil \checkmark \\ \hline

\end{tabular}
\end{table}

\section{Fairness in Semi-Supervised Learning} \label{section: appendix SSL}

SSL is a type of ML that falls in between supervised and unsupervised algorithms. It allows the use of both labeled and unlabeled data, which is crucial due to the tremendous need for data in large models and the high cost of generating labeled data. Moreover, previous research has shown that more data leads to a better balance between fairness and accuracy. Despite the essential role of SSL and a handful of surveys on this topic, including~\cite{prakash2014survey, zhu2005semi, pise2008survey, van2020survey}, there exists limited work on understanding fairness in SSL. To our knowledge, no survey covers the implications of fairness in SSL. The existing SSL methods can be categorized into the following $4$ groups:

\begin{itemize}
    \item {\bf Wrapper Methods.} The objective of this group, with the most widely adopted algorithm being pseudo-labeling, is to train a classifier using the labeled data available and then forecast labels for unlabeled data points.
    \item {\bf Unsupervised Extraction Methods.} This strategy aims to obtain beneficial features from data that has not been labeled by clustering and to utilize the information gained to improve training on labeled data.
    \item {\bf Intrinsic Methods.} In this class, unlabelled data are directly incorporated into the objective function of learning methods commonly used in supervised objectives for labeled data.
    \item {\bf Graph-based Approaches.} In this method, a graph is constructed, where each data point serves as a node, and the similarity matrix reflects the connections and associations among them. This graph is then utilized to make predictions about nodes or labels.
\end{itemize}

The amalgamation of fairness concepts from supervised and unsupervised learning is seen in SSL, due to its interdisciplinary nature. Consequently, our primary focus is on the mitigation algorithms.

\subsection{Unfairness Mitigation Algorithms}

\subsubsection{Pre-processing Strategies} The approaches commonly employed as pre-processing techniques in SSL have been classified into two categories: fair embeddings and reweighting.

{\bf Fair Embedding.} Graph embedding aims to transform graph data into a low-dimensional vector space, where nodes are represented as vectors. This process helps to capture the underlying structure and relationships in data such that it can be used for SSL tasks such as node classification and link prediction. Learning fair embeddings in the pre-processing phase allows for a significant reduction of bias in subsequent SSL models.

Based on the node2vec algorithm~\cite{grover2016node2vec}, the authors in~\cite{rahman2019fairwalk} propose a method called Fairwalk for generating fair embeddings predicated on statistical parity. In Fairwalk, instead of randomly selecting a node to jump to from all of its neighbors, neighbors are divided into clusters based on their sensitive attribute values, with each cluster having an equal probability of being chosen, regardless of its size. Then, a node is randomly selected from the chosen cluster for the jump. Meanwhile, in another recent paper, Fan et al.~\cite{fan2021fair} introduced FairGAE, a method that employs an auto-encoder model to generate unbiased graph embeddings. FairGAE achieves the objective by blocking the message-passing procedure from certain neighbors of nodes based on their sensitive attributes, so that each node has an equal chance of being influenced by other groups. By doing so, FairGAE ensures that the graph convolutional network mechanism depends only on the network structure, rather than on sensitive attributes.

{\bf Reweighting.} The reweighting technique tends to boost the associated weight of certain groups or individuals in order to improve fairness in SSL outcomes. Khajehnejad et al.~\cite{khajehnejad2022crosswalk} proposed a method to reweight graph edges in order to promote fairness in random walks. Specifically, edges that connect different groups or are in close proximity to group boundaries are given greater weight. This approach leads to more transitions across group boundaries during the random walk process, which results in graph embeddings that better capture the structure of the entire network. The authors also demonstrate that this approach improves the performance of SSL algorithms in terms of statistical parity.\\

\subsubsection{In-processing Strategies} We have grouped in-processing techniques into two broad categories of {\em adversarial regularizers} and {\em fairness regularizers}.

{\bf Adversarial Regularizers}. In this group, a discriminator is trained to learn sensitive attributes during training. Such information is then commonly added as an extra regularizer in the loss function. The majority of adversarial regularizers in SSL are explored for node classification in GNNs. As proven in~\cite{li2018deeper, wang2020unifying}, embedding nodes with connected components will be closer even after one aggregation of message-passing. Therefore, it is understandable nodes with similar sensitive attributes, once used together, tend to reach similar embedding. The authors in~\cite{dai2021say} demonstrate this susceptibility of GNNs on different GNN architectures and propose a framework called FairGNN to address it. The framework utilizes an estimator to learn sensitive attributes of nodes and incorporates an adversarial learner in GNN that ensures that node predictions are independent of sensitive attributes. Additionally, fairness regularizers based on statistical parity and equalized odds are used in the objective to achieve group fairness. The authors in~\cite{bose2019compositional} focus on how node embedding can be learned such that they do not correlate with sensitive attributes. The mechanism to achieve fair embeddings is by introducing a set of adversarial filters applied to remove information about sensitive attributes. The filters are learned during the training and could also be used as a postprocessing approach to ensure embeddings are invariant with respect to protected features.

{\bf Fairness Regularizers} In this group, fairness constraints are implemented directly into the loss function without requiring a discriminator. The method in~\cite{wang2022unbiased} assumes that a bias-free graph can be generated from pre-defined non-sensitive attributes. The non-sensitive attributes are used to construct a graph that is assumed to be free of bias. The authors then propose a regularization term that encourages the learned embeddings to satisfy certain fairness properties that are consistent with the bias-free graph. The regularization term is designed to penalize differences between the learned embeddings and the embeddings that would be obtained from the bias-free graph. The framework in~\cite{agarwal2021towards} considers counterfactual fairness and aims to maximize the similarity between representations of the original nodes in the graph, and their counterparts in the augmented graph. This is done by the introduction of a new learning objective function. Each counterfactual example is generated by modifying sensitive attributes and random masking of sensitive attributes.\\

\subsubsection{Post-processing Strategies} Several techniques outlined for both supervised and unsupervised learning can be employed as post-processing tactics in SSL. However, we have recognized "Fair Boosting" as the principal approach primarily utilized in SSL.

{\bf Fair Boosting.} Iosifidis et al.~\cite{iosifidis2019adafair} propose an algorithm termed AdaFair that builds upon the AdaBoost algorithm to enhance fairness during Boost rounds. As opposed to the AdaBoost, where in each round weak classifier only takes into account the hard classification, AdaFair additionally considers the discriminated group as they are dynamically being identified. Then, follows a reweighting strategy based on the accumulative fairness notion to tackle the problem of class imbalance. Zhu et al.~\cite{zhu2021rich} focus on pseudo labeling and reveal the disparate impact in SSL. The authors demonstrate that subpopulations with higher baseline accuracy levels without SSL benefit more from SSL. The opposite is also true, meaning subgroups who suffer low baseline accuracy tend to experience performance drop after SSL. An evaluation metric called Benefit Ratio is proposed to capture the normalized accuracy improvement on subgroups. The authors in~\cite{zhang2020fairness} propose a post-processing approach on top of pseudo labeling to lower discrimination-level explained in the following three steps. First, pseudo labeling is conducted based on a relatively small portion of data that are labeled. Second, several training datasets are generated by sampling from the pseudo labeled dataset. The sampling is conducted such that groups are fairly represented in terms of statistical parity. Finally, separate models are trained over the fair datasets, and ensemble learning is used to select the labels with the highest vote.

\begin{spacing}{0.9}
\bibliographystyle{ACM-Reference-Format}
\bibliography{Ref_main,Ref_privacy}
\end{spacing}


\end{document}